\pdfoutput=1

\documentclass[11pt]{article}
\usepackage[final]{acl}

\usepackage{times}
\usepackage{latexsym}
\usepackage[T1]{fontenc}

\usepackage[utf8]{inputenc}
\usepackage{microtype}
\usepackage{inconsolata}
\usepackage{array}  

\usepackage{graphicx}
\usepackage{multirow}
\usepackage{hyperref}
\usepackage{makecell}  
\usepackage{colortbl}
\usepackage{algorithm}
\usepackage{algorithmic}
\usepackage{amssymb}

\title{Problem-Solving Logic Guided Curriculum In-Context Learning \\for LLMs Complex Reasoning}

\author{Xuetao Ma, Wenbin Jiang, Hua Huang\thanks{~Corresponding author: Hua Huang} \\
        School of Artificial Intelligence, Beijing Normal University, Beijing, China \\
        \texttt{maxuetao@mail.bnu.edu.cn}, \texttt{\{jiangwenbin, huahuang\}@bnu.edu.cn}
        }
\begin{document}
\maketitle

\begin{abstract}
    In-context learning (ICL) can significantly enhance the complex reasoning capabilities of large language models (LLMs), with the key lying in the selection and ordering of demonstration examples.
    Previous methods typically relied on simple features to measure the relevance between examples.
    We argue that these features are not sufficient to reflect the intrinsic connections between examples.
    In this study, we propose a curriculum ICL strategy guided by problem-solving logic.
    We select demonstration examples by analyzing the problem-solving logic and order them based on curriculum learning.
    Specifically, we constructed a problem-solving logic instruction set based on the BREAK dataset and fine-tuned a language model to analyze the problem-solving logic of examples.
    Subsequently, we selected appropriate demonstration examples based on problem-solving logic and assessed their difficulty according to the number of problem-solving steps.
    In accordance with the principles of curriculum learning, we ordered the examples from easy to hard to serve as contextual prompts.
    Experimental results on multiple benchmarks indicate that our method outperforms previous ICL approaches in terms of performance and efficiency, effectively enhancing the complex reasoning capabilities of LLMs.
    Our project will be released at \url{https://github.com/maxuetao/CurriculumICL}
\end{abstract}

\section{Introduction}

\begin{figure}[!ht]
  \centering
  \includegraphics[width=0.9\linewidth]{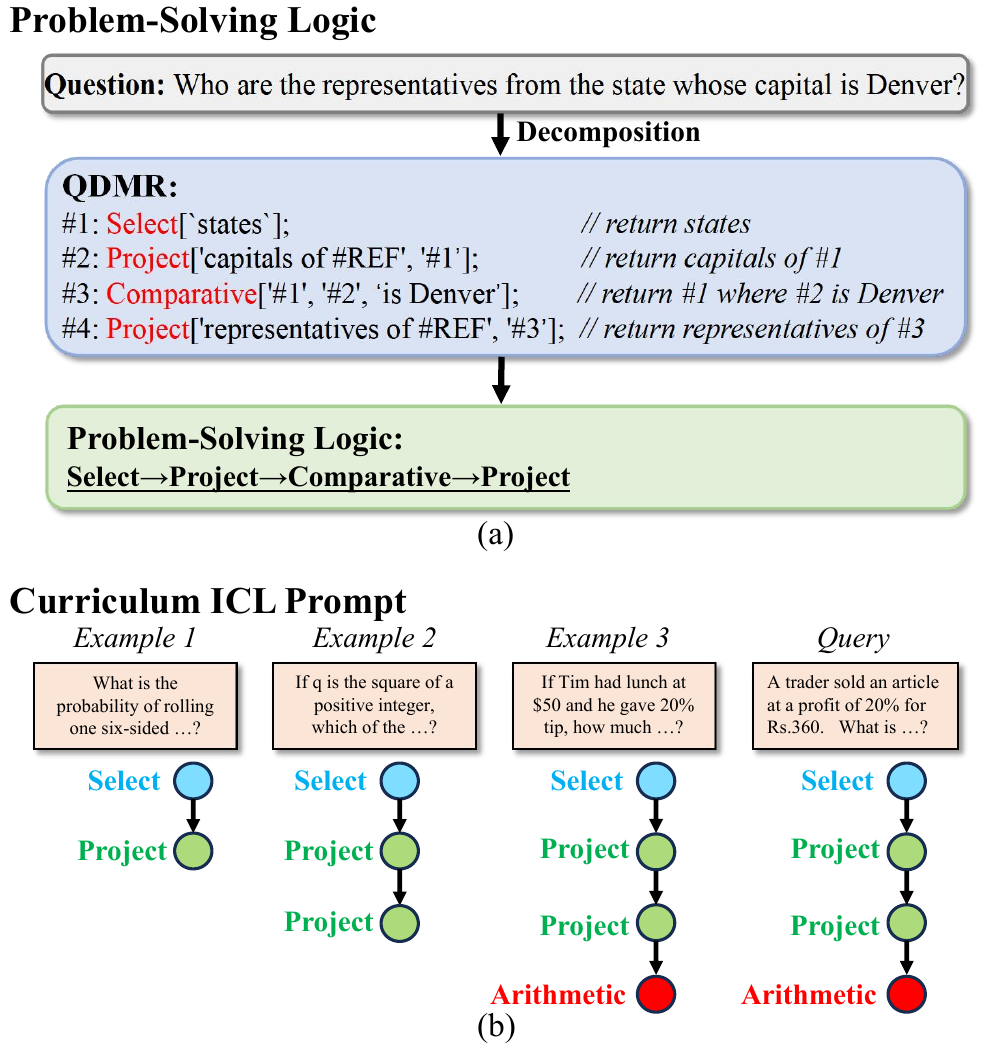}
  \caption{(a) The transformation from QDMR to problem-solving logic. (b) An example of curriculum ICL. Example selection depends on the similar problem-solving logic, and example ordering depends on the number of operations contained in the logic.
  }\label{psl_and_icl}
\end{figure}

Large language models (LLMs)~\cite{human_tuning,gpt3,chatgpt} can rapidly acquire new capabilities through in-context learning (ICL) to solve many new tasks~\cite{iclsurvey1,iclsurvey2}, and can be extended through chain of thought (CoT)~\cite{cot} to solve many tasks that require complex reasoning~\cite{complex2,complex3,complex4}.
Researchers believe that through ICL, LLMs can implicitly learn the problem-solving patterns demonstrated in contextual examples and apply them to new tasks~\cite{understandicl,understandicl2}.
This means that LLMs have the ability to learn and apply problem-solving patterns on the spot from given examples.

In recent years, supervised fine-tuning (SFT) methods~\cite{sft1,sft2} and reinforcement learning methods~\cite{rl1,deepseek,rl2} have been able to significantly enhance the abilities of LLMs through training.
Despite this, due to the unique characteristic of ICL that it can enhance problem-solving capabilities without training, it still holds value as significant as the methods mentioned above,
especially when facing the need to reduce costs or quickly apply to new tasks.
Relevant work~\cite{smallsft1} has already shown that LLMs possess a wealth of basic knowledge and fundamental capabilities that can be effectively activated through a small number of examples.
Particularly, LIMO~\cite{lessismore} fine-tuned a large language model with only a few hundred examples and achieved results that are close to or even on par with the current state-of-the-art reinforcement learning optimization inference.
Therefore, we believe that the ICL capabilities of current LLMs are still far from being fully realized.
There is a need to design better prompts to effectively enhance the effectiveness of ICL.

ICL learns demonstration examples in sequence and then solves problems, which closely resembles the process of humans learning knowledge step by step.
We believe that organizing demonstration examples in a way similar to human educational curriculum construction is crucial.
It helps LLMs learn the knowledge and patterns shown in the examples and solve given problems effectively.
Therefore, strategies for curriculum learning~\cite{curriculum} can be adopted for the organization of demonstration examples.
The key to ICL lies in example selection and ordering, which requires measuring the relevance between examples.
Traditional simple statistical information, such as similarity~\cite{similarity,similarity2,similarity3} and perplexity~\cite{perplexity,perplexity2}, is not sufficient to reflect the intrinsic connections between examples, especially from the perspective of problem-solving.

In this work, we innovatively propose a problem-solving logic guided curriculum ICL method, which constructs the optimal ICL prompt for the query based on problem-solving logic.
The Question Decomposition Meaning Representation (QDMR)~\cite{qdmr} decomposes complex problems into several sub-questions for solving and formalizes these sub-questions with 13 custom "operations", which we refer to as \textit{problem-solving logic}.
Figure~\ref{psl_and_icl}-(a) shows an example of problem decomposition and transformation into problem-solving logic.
Although it cannot directly solve the problem, the problem-solving logic describes the steps required for solving and the order of these steps in formal language.
Therefore, it can accurately measure the intrinsic connections between examples and construct a sequence of demonstration examples that are conducive to problem-solving.
Figure~\ref{psl_and_icl}-(b) shows an example of curriculum ICL.
We select examples with similar problem-solving logic, which can help LLMs learn how to solve similar problems.
Subsequently, we measure the difficulty of these examples by the number of problem-solving steps.
The greater the number of steps, the more reasoning steps are involved, meaning the problem is more difficult to solve.
Relying on the principles of curriculum learning, we order these examples from easy to hard to serve as the final in-context prompt.

Our main contributions are as follows:

(1) This paper proposes a problem-solving logic guided curriculum ICL strategy to enhance the reasoning performance of LLMs.
We innovatively present problem-solving logic as the criterion for selection and ordering demonstration examples, which is expected to offer a novel perspective for future work.

(2) We constructed a problem-solving logic instruction set based on the BREAK dataset.
Based on this, we fine-tuned a language model to automatically analyze the problem-solving logic of input questions.

(3) Extensive experiments are conducted on five datasets, and results show that our method achieves significant improvements in average performance and efficiency across all datasets, surpassing previous ICL methods and effectively enhancing the ability of LLMs in reasoning tasks.

\section{Background}
\subsection{In-Context Learning}
ICL is a capability that emerges as the training data and scale of LLMs increase~\cite{defineicl}.
This allows LLMs to learn new tasks with only a few examples.
Examples generally contain questions and answers.
The query needs to maintain consistent formatting with the examples so that LLMs can provide accurate responses.
This process is called few-shot.

Existing research shows that the key to enhancing ICL performance lies in the organization of demonstration examples, that is, the selection and ordering of examples.
Taking text similarity as an example, the general process is to encode the candidate examples and the query into vector forms, and then select the examples most similar to the query by calculating the similarity between vectors.
Subsequently, these examples are sorted according to text similarity.
Finally, the sorted examples are then input into the LLMs together with the query for solving.

\begin{figure}[!ht]
  \centering
  \includegraphics[width=1\linewidth]{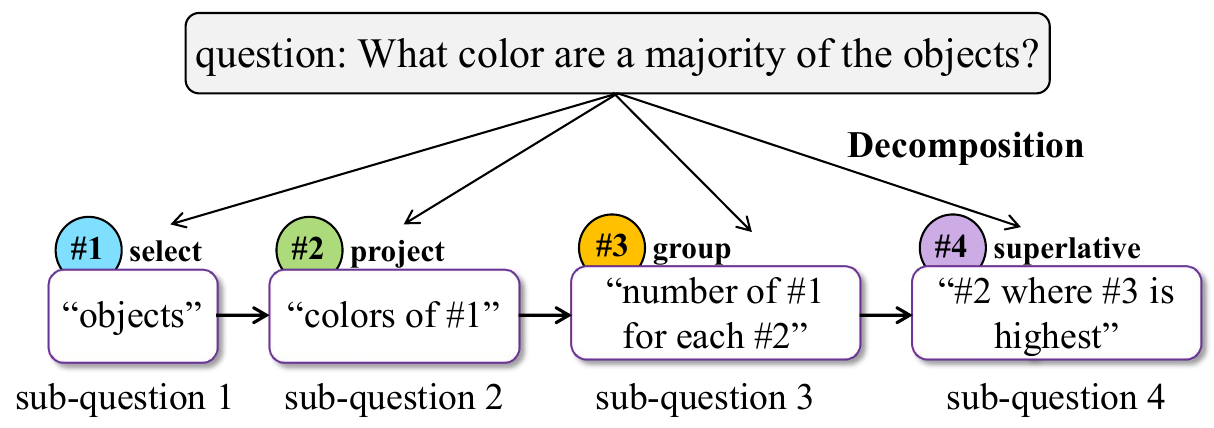}
  \caption{A QDMR example. The original question is decomposed into four sub-questions, each represented by an operation.}
  \label{qdmr}
\end{figure}

\subsection{Problem-Solving Logic}
QDMR is a general method for decomposing complex questions into several sub-questions for solving.
They manually designed 13 operations, with each sub-question represented by an operation.
The researchers proposed the BREAK dataset through manual annotation, which contains 60K question-answer pairs.
Specific examples of each operation, as well as detailed information about the dataset, can be found in the Appendix~\ref{append1}.

This work is inspired by QDMR and refers to the sequence of operators representing sub-questions as the \textit{problem-solving logic}.
The set of sub-questions decomposed by QDMR includes the required steps and the order between steps.
Figure~\ref{qdmr} shows a specific QDMR   example.
The original question is split into four sub-questions, each of which is described in a formal language with an operation, resulting in the corresponding problem-solving logic as follows:
\[
\mathrm{select} \rightarrow \mathrm{project} \rightarrow \mathrm{group} \rightarrow \mathrm{superlative}
\]

\subsection{Curriculum Learning}
Curriculum learning is a machine learning strategy~\cite{curriculum}.
It suggests that the training process should mimic human cognitive learning by starting with simple examples and gradually increasing in difficulty.
The core of this method lies in how to measure the difficulty of examples, which often depends on the characteristics of the specific task.
For example, in the field of computer vision, the number of objects in an image~\cite{image_curriculum} or noise~\cite{noise_curriculum} contained can be used to measure difficulty.
In the field of natural language processing, sentence length~\cite{textlength_curriculum} can be used as a measure of difficulty.
In addition to these, the difficulty can also be measured by human educational level~\cite{humancurriculum} or evaluation models~\cite{score_curriculum}.

\begin{figure*}[!ht]
  \centering
  \includegraphics[width=0.9\textwidth]{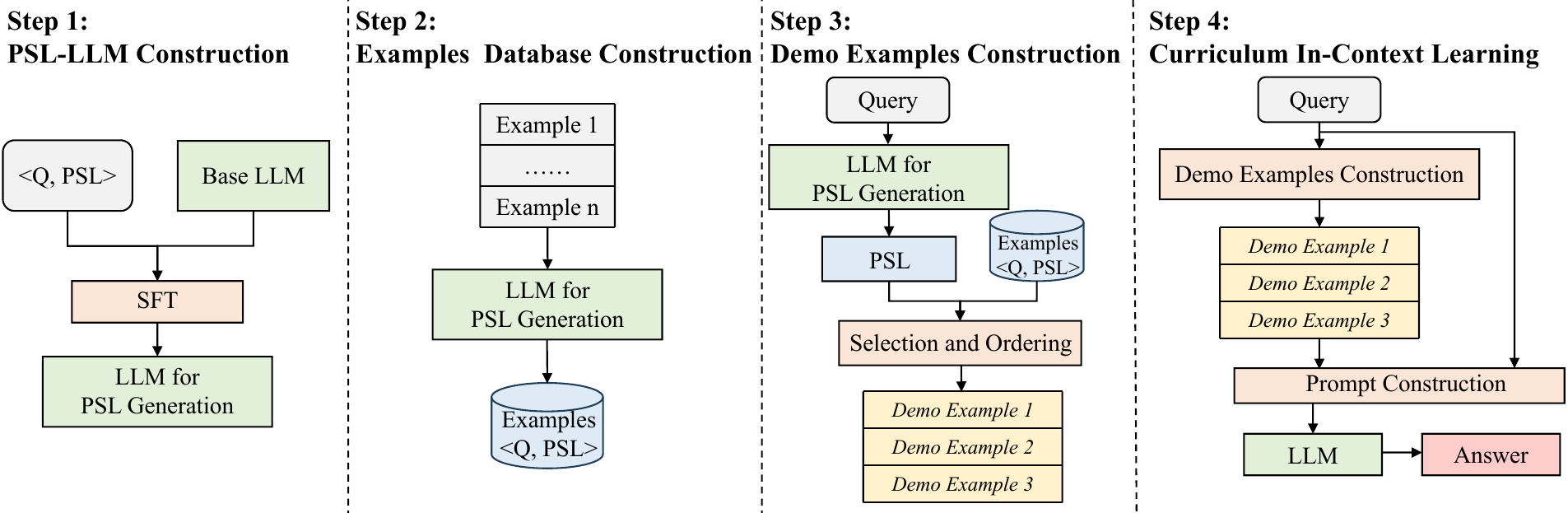}
  \caption{The overall flowchart of our method. First, a base LLM is fine-tuned using an instruction set for problem-solving logic (PSL) constructed from the BREAK dataset. Then, suitable demonstration examples are selected and ordered by analyzing the PSL of the candidate examples and the query. Finally, the selected demonstration examples and the query form the full prompt, which is fed into the LLM to obtain the results.}
\label{workflow}
\end{figure*}
\section{Problem-Solving Logic Guided Curriculum ICL}
This paper introduces a problem-solving logic guided curriculum ICL strategy.
The overall methodology is illustrated in Figure~\ref{workflow}.
Specifically, we first constructed an instruction set based on the BREAK dataset and fine-tuned a language model to automatically analyze problem-solving logic.
Then, we analyzed the problem-solving logic for all data in the benchmark training set to construct a dataset of candidate examples.
When an actual query is input, its problem-solving logic is first analyzed and then compared with the candidate examples, selecting those with similar problem-solving steps as demonstration examples.
Furthermore, the number of problem-solving steps serves as an appropriate metric for assessing the difficulty of each example.
A greater number of steps means the problem is more difficult to solve.
This inspired us to apply the principles of curriculum learning to order the demonstration examples from easy to hard.
Finally, the ordered demonstration examples and the query are combined to form the final prompt, which is then input into the LLMs.
The following sections will offer a detailed explanation of how problem-solving logic is analyzed, along with the process of selecting and ordering demonstration examples.

\subsection{Problem-Solving Logic Analysis}
We first need to train a language model to analyze the problem-solving logic, which is represented as an ordered set of several problem-solving steps.

Our approach constructs an instruction set based on the BREAK dataset.
Specifically, the input to the instruction set is a problem, and the output is problem-solving logic and its formal language.
The formal language ensures that the model correctly understands the problem-solving process.
We then fine-tune a Llama3-8B model~\cite{llama2, llama3} with LoRA~\cite{lora} on this instruction set.
Once the model is trained, it can analyze problems from any dataset and extract their problem-solving logic.
Examples of the instruction set can be found in the Appendix~\ref{append1}. Details of fine-tuning and hyperparameters can be found in the Appendix~\ref{finetune}.

Analyzing the problem-solving logic is a crucial step in our work, providing the foundation for the subsequent curriculum ICL.

\begin{figure}[!ht]
  \centering
  \includegraphics[width=0.9\linewidth]{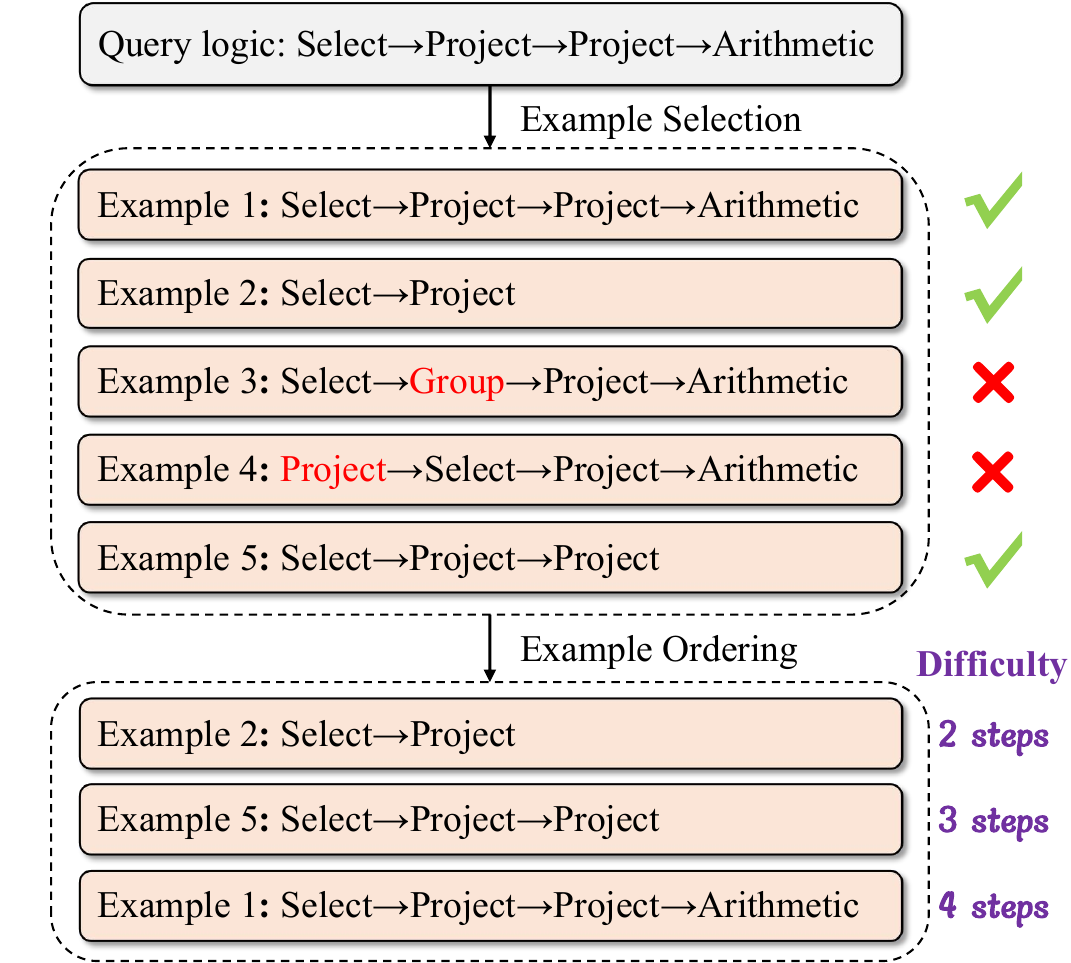}
  \caption{
    The process of example selection and ordering.
    $(\checkmark)$ denotes similar problem-solving logic, $(\times)$ indicates a matching failure, and \textcolor{red}{\textbf{red}} font indicates the reason for the matching failure.
    \textcolor[RGB]{125,93,149}{\textbf{Difficulty}} is measured by the number of steps.
  }
  \label{selection_and_ordering}
\end{figure}

\subsection{Curriculum ICL}
Based on the above problem analysis process, we can focus on problem-solving logic to guide the selection and ordering of demonstration examples.
Figure~\ref{selection_and_ordering} illustrates the process of example selection and ordering.

\subsubsection{Demonstration Example Selection}
First, we need to select appropriate demonstration examples.
Compared to semantic information, we believe that selecting examples with similar problem-solving logic is more important.
On one hand, similar problem-solving logic can guide LLMs in reasoning, and on the other hand, examples with similar logic but different semantics can enhance the model's generalization ability.

\begin{algorithm}[!ht]
  \caption{Demonstration Example Selection}
  \label{selection}
  \begin{algorithmic}[1]
  \REQUIRE query $T$, LLM function $F(\cdot)$, set of candidate examples $\{E_1, E_2, \dots, E_n\}$, each example $E_i$ has its own solution logic $L_i = \{O_{i1}, O_{i2}, \dots, O_{im_i}\}$.
  \ENSURE Mark matching demonstration examples.
  \STATE $L_T \gets F(T)$ \COMMENT{Obtain the solution logic for the query from LLM}
  \FOR{each example $E_i$ in $\{E_1, E_2, \dots, E_n\}$}
      \STATE $L_i \gets \{O_{i1}, O_{i2}, \dots, O_{im_i}\}$ \COMMENT{Retrieve solution logic of $E_i$}
      \IF{$L_i$ is a subsequence of $L_T$ starting from the first operator}
          \STATE Mark $E_i$ as a demonstration example
      \ENDIF
  \ENDFOR
  \end{algorithmic}
\end{algorithm}

\begin{figure*}[!ht]
  \centering
  \includegraphics[width=0.9\textwidth]{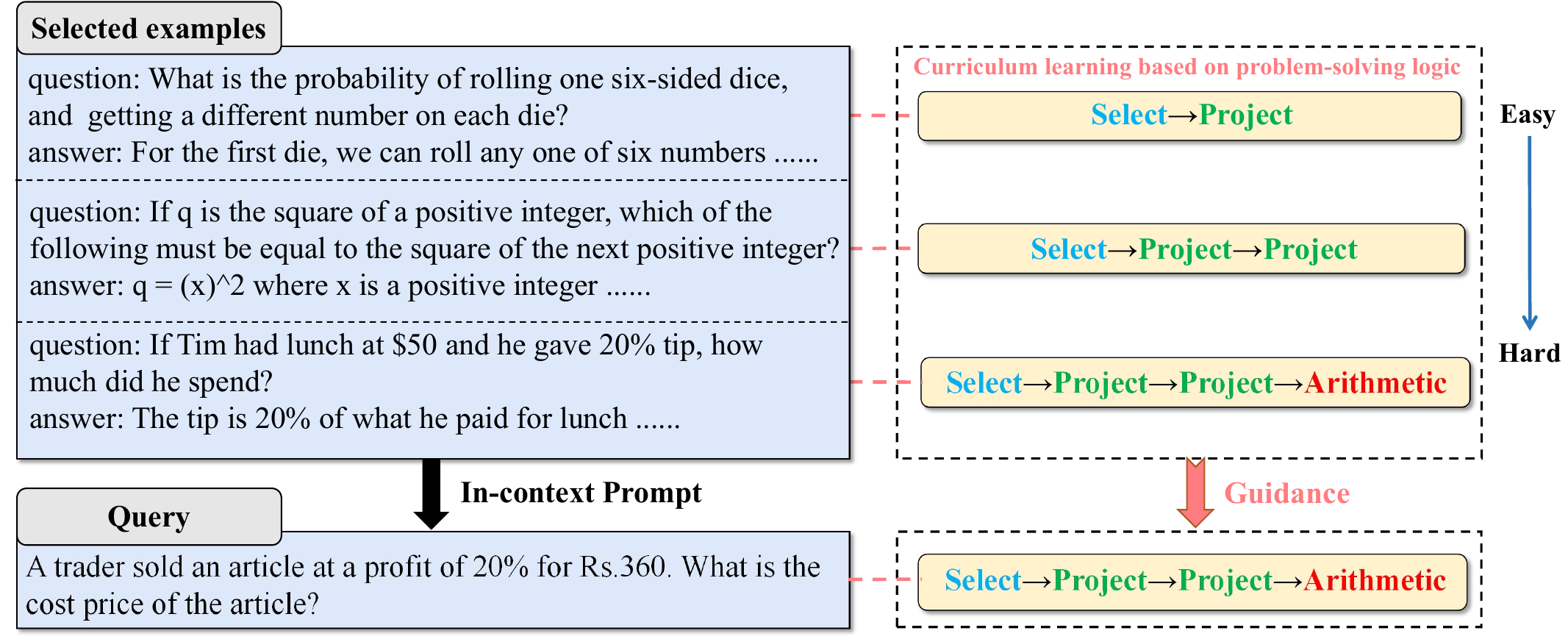}
  \caption{
    A complete example of curriculum ICL. The selected examples form the context information. The right half of the figure shows the problem-solving logic, which is the basis for example selection and ordering.}
  \label{icl}
\end{figure*}

After analyzing the query and all candidate examples, our method selects demonstration examples based on the problem-solving logic.
The selection criterion requires that the problem-solving operations set in each candidate example must be a subsequence of the query, meaning both the types of operations and their order must match exactly.
Suppose the query has a problem-solving logic containing $m$ operations, and the selected demonstration example has $n$ operations $(m \geq n)$; the $n$ operations of the demonstration example must match the first $n$ operations of the query. 
This method ensures that the demonstration example's problem-solving steps align with the first $n$ steps of the query, avoiding any mismatch or additional problem-solving steps.
The complete process is detailed in Algorithm~\ref{selection}.

\subsubsection{Demonstration Examples Ordering}
The key to curriculum learning lies in how to measure the difficulty of examples.
By introducing problem-solving logic, we can easily assess the difficulty of each example.
The problem-solving logic consists of several operations, where a higher number of operations indicates more reasoning steps, thereby increasing the problem's difficulty.

Inspired by this, we applied curriculum learning principles, ordering examples from easy to hard.
Specifically, we sorted the examples in increasing order based on the number of problem-solving steps, and used them along with the query to construct the final in-context prompt.
Figure~\ref{icl} shows a complete curriculum ICL example, including demonstration examples and the query.

  \begin{table*}[!ht]
    \begin{small}
    \centering
    \begin{tabular}{ccccccccc}
    \hline
    \multirow{2}{*}{\textbf{Method}} & \multirow{2}{*}{\makecell{\textbf{Selection} \\ \textbf{Stategy}}} & \multirow{2}{*}{\makecell{\textbf{Ordering} \\ \textbf{Stategy}}} & \multicolumn{5}{c}{\textbf{Dataset}} & \multirow{2}{*}{\textbf{Avg.}} \\ 
    \cline{4-8}
    &  & & \textbf{SVAMP}  &\textbf{AQuA} & \textbf{Gsm8k} & \textbf{ComSenQA} & \textbf{StrategyQA} & \\ \hline
    \rowcolor{gray!20} \textbf{\textit{Llama3-8B}} &  &  &  &  &  & & & \\
    Random       & Random   &  Random  &76.5  & 46.5  & 73.8  & 75.8  & 65.1 & 67.53  \\
    
    VoteK       & KNN &  Similarity  &74.9  & 44.9  & 76.7  & 75.4  & 69.0 & 68.19  \\
    
    PromptSO      & PCA   &  Eigenvalue  & 77.3  & 43.7  & 77.7  & 75.6  & 67.7 & 68.40  \\
    
    AutoCoT       & K-means  & Similarity &77.5  & 47.2  & 75.3  & 76.0  & \underline{71.2} & 69.44  \\
    
    CoT + Fewshot & Fixed    &  Fixed  &\underline{80.5}  & 44.5 & \underline{79.4}  & 75.1  & 69.4 & 69.79  \\
    
    SA-ICL      &  KNN  &  Entropy  &78.8  & \underline{47.6}  & 77.9  & \textbf{78.5}  & 66.8 & 69.95  \\
    
    AL-ICL   & KNN & Similarity &80.8  & 45.7  & 78.2  & \underline{77.9}  & 68.1 & \underline{70.13}  \\
    Ours  & PSL  & Curriculum & \textbf{83.4} & \textbf{50.8} & \textbf{81.1} & 75.0 & \textbf{71.6} & \textbf{72.37}  \\
    
    \rowcolor{gray!20} \textbf{\textit{Llama3-70B}} &  &  &  &  &  & & & \\
    Random       & Random   &  Random  &85.9  & 69.7  & 91.6  & \underline{81.7}  & 80.3 & 81.84  \\
    
    VoteK       & KNN &  Similarity  &86.2  & 67.3  & 92.3  & 81.6  & 82.5 & 81.98  \\
    
    PromptSO      & PCA   &  Eigenvalue  & 86.8  & \textbf{73.6}  & 90.1  & 81.0  & 77.3 & 81.76  \\
    
    AutoCoT       & K-means  & Similarity & \underline{89.2}  & 65.0  & \underline{92.5}  & 81.4  & 74.2 & 80.46  \\
    
    CoT + Fewshot & Fixed    & Fixed  & 88.8  & 67.7 & 91.6  & 81.4  & 82.5 & 82.40  \\
    
    SA-ICL      &  KNN  &  Entropy  &85.7  & 67.3 & 92.3  & 81.1  & 82.5 & 81.78  \\
    
    AL-ICL   & KNN & Similarity &87.2  & 68.5  & 91.7  & 81.2  & \underline{84.2} & \underline{82.56}  \\
    Ours  & PSL  & Curriculum & \textbf{90.5} & \underline{70.1} & \textbf{92.6} & \textbf{81.8} & \textbf{85.2} & \textbf{84.04}  \\

    \rowcolor{gray!20} \textbf{\textit{Qwen2.5-7B}} &  &  &  &  &  & & & \\
    Random       & Random   &  Random  &87.3  & 74.8  & 87.2  & 83.4  & 65.0 & 79.54  \\
    
    VoteK       & KNN &  Similarity  & 86.2  & \textbf{79.9}  & 84.6  & 83.5  & 68.1 & 80.46  \\
    
    PromptSO      & PCA   &  Eigenvalue  & 85.6  & 76.4  & \textbf{90.5}  & 82.7  & \underline{70.3} & 81.11  \\
    
    AutoCoT       & K-means  & Similarity & 87.1  & 69.7  & \underline{90.4}  & 83.6 & 69.7 & 80.10  \\
    
    CoT + Fewshot & Fixed    &  Fixed  & \underline{90.2}  & 76.4 & 89.8  & \underline{83.8} & 63.8 & 80.79  \\
    
    SA-ICL      &  KNN  &  Entropy  & 87.7 & 71.7 & 88.7  & 83.3 & 69.0 & 80.07 \\
    
    AL-ICL   & KNN & Similarity & 88.4 & 77.6  & 89.5 & 82.5  & 67.7 & \underline{81.13}  \\
    Ours  & PSL  & Curriculum & \textbf{92.3} & \underline{78.3} & 90.0 & \textbf{84.6} & \textbf{71.2} & \textbf{83.28}  \\

    \hline
    \end{tabular}
    \caption{\label{table1}
    The table presents a comparison of experimental results across different benchmarks using Llama3-8B, Llama3-70B and Qwen2.5-7B, demonstrating the accuracy contrast between various ICL methods.
    \textbf{Avg} represents the average accuracy across the different benchmarks.
    The best and second-best performances are highlighted in \textbf{bold} and \underline{underlined}, respectively.}
    \end{small}
  \end{table*}
  
  \section{Experiments and Analysis}
  
  \subsection{Experimental Setup}
  \paragraph{Benchmarks.}
  Our experiment includes two types of datasets, Arithmetic Reasoning and Commonsense Reasoning, and validation is conducted on five different datasets.
  Arithmetic Reasoning: (1) the AQuA \cite{aqua} includes 254 test examples, (2) the SVAMP \cite{svamp} includes 1000 test examples, (3) the Gsm8k includes 1319 test examples. 
  Commonsense Reasoning: (1) the CommonsenseQA \cite{commonsenseqa} includes 1211 test examples, (2) the StrategyQA \cite{strategyqa} includes 229 test examples.

  \paragraph{Baselines.}
  We compare our approach against seven methods that use ICL.
  Random selects demonstration examples and their order randomly.
  VoteK \cite{votek} selects the most similar k examples using k-nearest neighbors (KNN) and sorts them according to similarity scores.
  PromptSO \cite{promptso} uses principal component analysis~\cite{pca} to select the most relevant basis questions and sorts them based on eigenvalue.
  AutoCoT \cite{autocot} uses k-means to automatically select the most representative examples that are closest to the cluster center.
  CoT+few-shot \cite{cot} manually designed fixed demonstration examples with reasoning processes.
  Self-Adaption ICL (SA-ICL)~\cite{saicl} selects similar examples based on KNN and then chooses an appropriate order based on information compression.
  Active Learning ICL (AL-ICL) \cite{alicl} selects most similar examples based on the principles of active learning and sorts them according to similarity.

  \paragraph{Implement Details.}
  We evaluate the effectiveness of our method on the Llama3-8B, Llama3-70B and Qwen2.5-7B~\cite{qwen2_5}. For each benchmark, we select demonstration examples from its training set to form prompt information to evaluate each test set data.
  For the SVAMP dataset, we adopted the same evaluation strategy as in previous work~\cite{svamp}, using ASDiv-a~\cite{asdiv} and MAWPS~\cite{mawps} together as the training set.
  To ensure a fair comparison, the number of selected examples for all baselines is based on the settings in CoT \cite{cot} for different benchmarks, and the number of examples selected by our method does not exceed that of the baselines.
  
  \subsection{Main Results and Analysis}
  We compare the performance of our approach with other ICL methods. All the comparison results are tabulated in Table~\ref{table1}.
  For Llama3-8B, compared with other methods, we achieve the best performance on SVAMP, AQuA, Gsm8k and StrategyQA. The average accuracy across all benchmarks is improved by 2.24\%.
  For Llama3-70B, we achieve the best performance on SVAMP, Gsm8k, ComSenQA and StrategyQA. The average accuracy across all benchmarks is improved by 1.48\%.
  For Qwen2.5-7B, we achieve the best performance on SVAMP, ComSenQA and StrategyQA. The average accuracy across all benchmarks is improved by 2.15\%.
  Overall, our proposed method demonstrates excellent performance across different parameter scales and different models, which shows the effectiveness and generalizability of the method.
  
  To further demonstrate the effectiveness of the method, we have conducted validation and analysis of the key strategies of the method.
    
  \begin{figure*}[t]
    \includegraphics[width=0.45\linewidth]{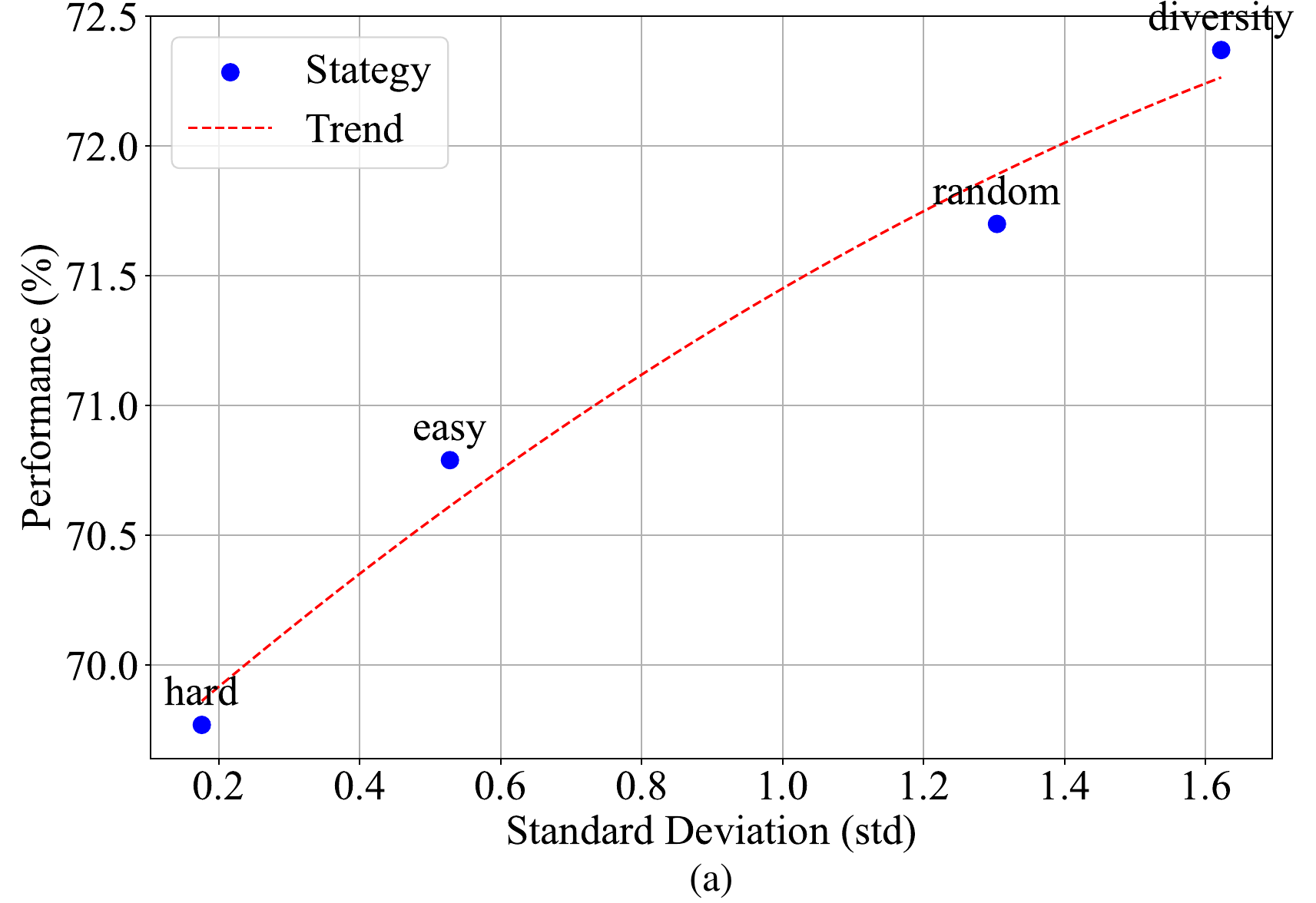} \hfill
    \includegraphics[width=0.45\linewidth]{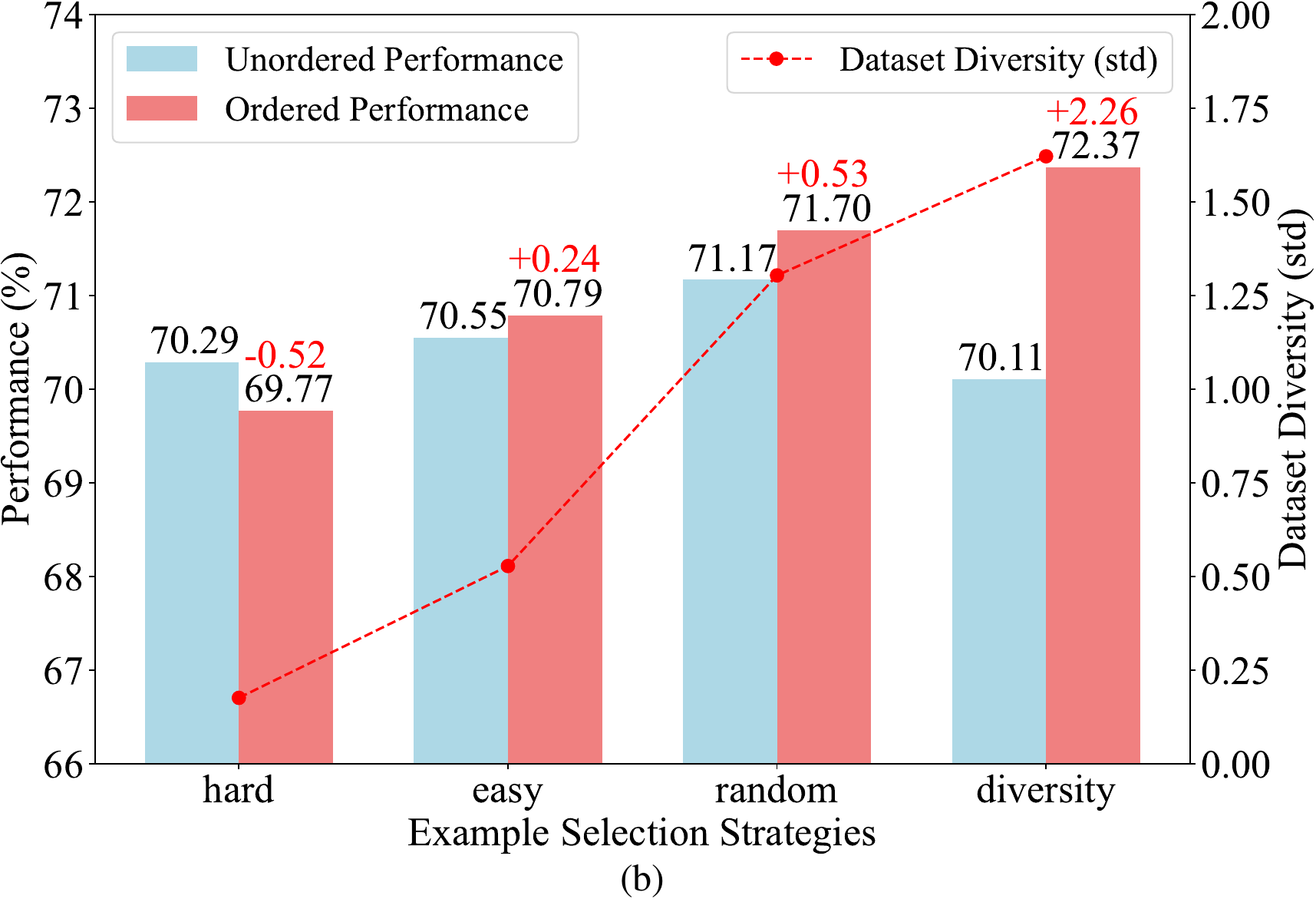}
    
  \caption{\label{analysis}
  (a) shows the relationship between the average standard deviation of different example selection strategies and their performance across various benchmarks.
  (b) shows the impact of example ordering strategies on performance in relation to the average standard deviation under different selection strategies.}
  \end{figure*}
  
  \begin{table*}[!ht]
    \begin{small}
    \centering
    \begin{tabular}{cccccccc}
    \hline
    
    \multirow{2}{*}{\makecell{\textbf{Difficulty} \\ \textbf{Strategy}}} & \multirow{2}{*}{\makecell{\textbf{Ordering}}} &\multicolumn{5}{c}{\textbf{Dataset}} & \multirow{2}{*}{\textbf{Avg.}} \\ 
    \cline{3-7}
    &  & \textbf{SVAMP}  &\textbf{AQuA} & \textbf{Gsm8k} & \textbf{ComSenQA} & \textbf{StrategyQA} & \\ \hline
    
    \rowcolor{gray!20} 
    \textbf{\textit{Original Llama}} &  &  &  &  &  & & \\
    
    AL-ICL   &   &80.8  & 45.7  & 78.2  & \textbf{77.9}  & 68.1 & 70.13  \\
    
    \rowcolor{gray!20} \textbf{\textit{Our Strategy}} &  &  &  &  &  & & \\
    Prioritize simplicity & w/ order &82.3 & 47.6 & 79.5 & 75.5 & 69.0 & 70.79  \\
                          & w/o order &\underline{82.5} & 47.2 & 78.8 & 76.1 & 68.1 & 70.55  \\
    
    Prioritize difficulty & w/ order &81.8 & 44.9 & 77.9 & 76.6 & 67.7 & 69.77  \\
                          & w/o order &81.6 & 46.1 & 79.6 & \underline{77.0} & 67.2 & 70.29  \\
    
    Select Randomly       & w/ order &81.3 & \underline{50.6} & \underline{80.2} & 76.1 & 70.3 & \underline{71.70}  \\
                          & w/o order &80.9 & 48.6 & 79.2 & 76.0 & \underline{71.2} & 71.17  \\
    
    Prioritize diversity  & w/ order &\textbf{83.4} & \textbf{50.8} & \textbf{81.1} & 75.0 & \textbf{71.6} & \textbf{72.37}  \\
                          & w/o order &80.5 & 46.1 & 80.1 & 76.0 & 65.9 & 70.11  \\
    
    \hline
    \end{tabular}
    \caption{\label{table2}
      The table presents the accuracy of benchmarks under different difficulty selection strategies. "w/ order" indicates that the examples are ordered based on curriculum learning, while "w/o order" means the examples are randomly ordered.
      The best and second-best performances are highlighted in \textbf{bold} and \underline{underlined}, respectively.
      }
    \end{small}
  \end{table*}
  
  \subsubsection{Analysis of Example Selection and Ordering}

  For the selection and ordering strategies of demonstration examples in ICL, we designed several sets of experiments to verify the effectiveness of our method.
  
  Regarding example selection, since each query may match far more examples than the specified limit during the problem-solving logic analysis, it is necessary to analyze specific difficulty sampling strategies.
  We designed four difficulty sampling strategies: 
  (1) \textbf{Prioritize simplicity}: This strategy selects easy examples first.
  (2) \textbf{Prioritize difficulty}: This strategy selects difficult examples first.
  (3) \textbf{Select randomly}: This strategy randomly selects examples of any difficulty.
  (4) \textbf{Prioritize diversity}: This strategy aims to select as many difficulty levels as possible, sampling at most one example from each difficulty level.

  Regarding the ordering of examples, to validate the effectiveness of curriculum learning, we designed two sets of controlled experiments.
  Under the four sampling strategies mentioned above, we applied two ordering strategy: (1) \textbf{difficulty increasing ordering (w/ order)} and (2) \textbf{random ordering (w/o order)}.
  
  We conducted experiments on Llama3-8B, and the results are shown in Table~\ref{table2}. Through analysis, we have made the following observations:
  
  First, it can be noted from the table that the performance of the strategies using the problem-solving logic and curriculum learning approach generally outperforms AL-ICL.
  The prioritize diversity (w/ order) strategy significantly outperforms the others, achieving an average accuracy of 72.37\%.
  
  Furthermore, the importance of curriculum learning is highlighted in our findings.
  For prioritize diversity strategies, the effect of ordering is particularly pronounced.
  In contrast, the impact of ordering is less significant for the prioritize simplicity and prioritize difficulty strategies.

  Based on the findings above and considering the characteristics of different selection strategies, we believe that the primary reason for these results is data diversity, or more specifically, difficulty diversity.
  To explain this phenomenon, we calculated the difficulty levels included in the demonstration examples for each data across all benchmarks and computed the average standard deviation.
  Standard deviation (std) is typically used to measure the degree of variation, and this metric helps illustrate the data diversity produced by different strategies.

  We analyzed two sets of data: first, the relationship between difficulty diversity and strategy performance; and second, the impact of difficulty diversity on the four strategies, considering both the cases with and without ordering.
  
  Figure~\ref{analysis}-(a) depicts the relationship between performance and difficulty diversity across the four selection strategies.
  There is a clear positive correlation between difficulty diversity and performance, suggesting that data diversity is key to improving performance.
  Additionally, Figure \ref{analysis}-(b) shows the relationship between the performance difference (with and without ordering) and difficulty diversity across the four selection strategies.
  We found that ordering strategies are highly sensitive to difficulty diversity.
  Overall, the higher the difficulty diversity, the greater the improvement brought by ordering. Notably, the prioritize diversity strategy saw the largest performance improvement with ordering.
  This highlights the effectiveness of curriculum learning, where it is essential to order data according to difficulty.
  At the same time, it supports the idea that measuring example difficulty by the number of problem-solving steps is a valid approach.
  
  \begin{table*}[!ht]
    \centering
        \begin{tabular}{cccccccc}
        \hline
        \multirow{2}{*}{\textbf{Strategy}} & \multicolumn{5}{c}{\textbf{Dataset}} & \multirow{2}{*}{\textbf{Avg.}} & \multirow{2}{*}{\textbf{Time}} \\ 
        \cline{2-6}
        & \textbf{SVAMP} & \textbf{AQuA} & \textbf{Gsm8k} & \textbf{ComSenQA} & \textbf{StrategyQA} & \\ \hline
        Fixed Examples        & 8 & 4 & 8 & 7 & 6 & 6.60 & 109\% \\
        Prioritize simplicity & 7.27 & 4 & 7.73 & 7 & 5.88 & 6.38 & 117\% \\
        Prioritize difficulty & 7.27 & 4 & 7.15 & 5.82 & 5.84 & 6.02 & 167\% \\
        Select Randomly       & 7.49 & 4 & 7.73 & 7 & 5.88 & 6.42 & 144\% \\
        \rowcolor[HTML]{FFCCC9}
        Prioritize diversity  & 2.16 & 3.19 & 3.38 & 3.19 & 1.8 & 2.74 & 100\% \\\hline
        \end{tabular}
    \caption{\label{table3}
    The number of demonstration examples selected by different selection strategies in benchmarks.~\textbf{Avg} represents the average number of demonstration examples selected for each data. \textbf{Time} indicates the time cost comparison across different strategies.
    The highlighted part represent the strategy with most efficient.}
    \end{table*}
  \subsubsection{Analysis of the Number of Examples}
  The number of demonstration examples for each query also has an important impact on the performance of ICL, as well as on the reasoning efficiency of LLMs.
  Table~\ref{table3} presents the number of demonstration examples included with each test data across different strategies.
  For comparison, we use the fixed number of examples in CoT~\cite{cot} as a reference.
  
  We find that the prioritize diversity strategy has significantly superior performance while also having the least average number of demonstration examples.
  The average number of demonstration examples for other strategies is more than 6, while priority diversity strategy only requires 2.74.
  Fewer examples indicate a shorter in-context length, which helps the reasoning speed of LLMs.
  Table~\ref{table3} also presents the average time cost under different strategies.
  We uses the priority diversity strategy as the baseline at 100\% to measure the time cost of other strategies.
  Experimental results show that, compared to other strategies, the prioritize diversity strategy has a time cost advantage, reducing consumption by 9\% to 67\%, effectively improving inference performance.
  
  Current studies have shown that an increase in the number of demonstration examples usually leads to improved performance~\cite{longicl}.
  Our method demonstrates that the quantity of examples is not the only influencing factor.
  This conclusion is consistent with numerous studies~\cite{diverse1,diverse2,revisiting}, which indicate that data diversity plays a critical role in enhancing the generalization capability of LLMs.

\section{Related Work}

\subsection{In-Context Learning}
GPT-3 \cite{firstfewshot} exhibited few-shot and zero-shot learning abilities during the pretraining phase.
CoT~\cite{cot} designed several fixed demonstration examples manually as in-context information, inspired further research on ICL~\cite{icl_future1,cot_future1,icl_future2}.

Subsequent research has shown that the key to ICL lies in demonstration examples selection and ordering~\cite{relatedwork_icl1,relatedwork_icl2,relatedwork_icl3}.
Regarding example selection, AutoCoT~\cite{autocot} used k-means clustering to select representative examples and leveraged zero-shot CoT to generate their reasoning process as demonstration examples.
PromptSO~\cite{promptso} used principal component analysis~\cite{pca} to encode text and calculate similarity to select examples.
Another work~\cite{promot_retrieve} points out that a retriever can be trained using annotated data to determine whether an example is suitable for a query.
Regarding example ordering, a study~\cite{order} randomly generated multiple combinations of example orderings to create probe sets.
By analyzing the entropy of predicted labels for each probe set, the researchers selected the best-performing order.
KATE~\cite{kate} explored ordering examples based on task relevance as well as length-based sorting.
Relevance-based ordering prioritizes examples closely related to the target task, while length-based sorting considers potential advantages for specific tasks.

\subsection{Curriculum Learning in LLMs}
Numerous applications across various fields have demonstrated that curriculum learning can effectively enhance model training outcome~\cite{curriculum2,curriculum_survey}.

Currently, some works have applied curriculum learning to LLMs~\cite{relatedwork_cl1, relatedwork_cl2}.
A common approach is to train the model with examples progressing from easy to hard during fine-tuning.
For instance, a study~\cite{humancurriculum} conducted fine-tuning on a structured dataset that strictly covers multiple educational stages to simulate the progressive learning characteristics of humans.
In the medical field, similarly, human-defined and automatically generated methods were used to annotate data difficulty, and LLMs in the medical question-answering domain were fine-tuned from easy to hard.~\cite{humancurriculum}.
Additionally, another work~\cite{datasetdecomposition} decomposed datasets into sequences of varying lengths, using sequence length as a metric to measure data difficulty.

Another common approach for applying curriculum learning to LLMs is ICL.
For example, ICCL~\cite{iccl} utilized human experts or LLM-driven metrics to assess data difficulty, and gradually increased the difficulty of demonstration examples from easy to hard.

\section{Conclusion}
This paper proposes a problem-solving logic guided ICL strategy. 
By analyzing the problem-solving logic, we measure the similarity between problems and select demonstration examples.
Additionally, the difficulty of problems is assessed based on the number of problem-solving steps, and the selected examples are ordered from easy to hard following the principles of curriculum learning.
Experimental results across multiple benchmarks demonstrate that our proposed method outperforms other ICL methods in terms of average performance, significantly improving the reasoning capabilities of LLMs.

\section*{Limitations}
Although our work improves the performance and efficiency of LLMs in reasoning tasks, there are still limitations for improvement.
First, due to hardware resource constraints, we only conducted experiments on LLMs at the 8B scale, and further validation of our method is necessary on larger models, such as those at the 70B scale, to fully demonstrate its effectiveness.
On the other hand, we observed in many-shot studies~\cite{longicl} that a significant increase in the number of examples leads to substantial improvements in reasoning performance.
However, due to the limitations of benchmarks and hardware resources, we were unable to evaluate the effect of curriculum learning when applied to a large number of examples.
We believe that when both the quantity and quality of examples are ensured, reasoning performance can be further improved, which will be a focus of our future work.

\section*{Potential Risks}
Our work does not carry any obvious risks.

\section*{Acknowledgements}
This project was supported by the Natural Science Foundation of China (Project Number 62437001).

\bibliography{custom}

\begin{thebibliography}{68}
\providecommand{\natexlab}[1]{#1}

\bibitem[{Abdi and Williams(2010)}]{pca}
Herv{\'e} Abdi and Lynne~J Williams. 2010.
\newblock Principal component analysis.
\newblock \emph{Wiley interdisciplinary reviews: computational statistics}, 2(4):433--459.

\bibitem[{An et~al.(2023)An, Lin, Fu, Chen, Zheng, Lou, and Zhang}]{similarity3}
Shengnan An, Zeqi Lin, Qiang Fu, Bei Chen, Nanning Zheng, Jian-Guang Lou, and Dongmei Zhang. 2023.
\newblock How do in-context examples affect compositional generalization?
\newblock In \emph{Proceedings of the 61st Annual Meeting of the Association for Computational Linguistics (Volume 1: Long Papers)}, pages 11027--11052.

\bibitem[{Bahrini et~al.(2023)Bahrini, Khamoshifar, Abbasimehr, Riggs, Esmaeili, Majdabadkohne, and Pasehvar}]{chatgpt}
Aram Bahrini, Mohammadsadra Khamoshifar, Hossein Abbasimehr, Robert~J Riggs, Maryam Esmaeili, Rastin~Mastali Majdabadkohne, and Morteza Pasehvar. 2023.
\newblock Chatgpt: Applications, opportunities, and threats.
\newblock In \emph{2023 Systems and Information Engineering Design Symposium (SIEDS)}, pages 274--279. IEEE.

\bibitem[{Bengio et~al.(2009)Bengio, Louradour, Collobert, and Weston}]{curriculum}
Yoshua Bengio, J{\'e}r{\^o}me Louradour, Ronan Collobert, and Jason Weston. 2009.
\newblock Curriculum learning.
\newblock In \emph{Proceedings of the 26th annual international conference on machine learning}, pages 41--48.

\bibitem[{Bertsch et~al.(2024)Bertsch, Ivgi, Alon, Berant, Gormley, and Neubig}]{longicl}
Amanda Bertsch, Maor Ivgi, Uri Alon, Jonathan Berant, Matthew~R Gormley, and Graham Neubig. 2024.
\newblock In-context learning with long-context models: An in-depth exploration.
\newblock \emph{arXiv preprint arXiv:2405.00200}.

\bibitem[{Bhattamishra et~al.(2023)Bhattamishra, Patel, Blunsom, and Kanade}]{understandicl}
Satwik Bhattamishra, Arkil Patel, Phil Blunsom, and Varun Kanade. 2023.
\newblock Understanding in-context learning in transformers and llms by learning to learn discrete functions.
\newblock \emph{arXiv preprint arXiv:2310.03016}.

\bibitem[{Brown et~al.(2020)Brown, Mann, Ryder, Subbiah, Kaplan, Dhariwal, Neelakantan, Shyam, Sastry, Askell, Agarwal, Herbert-Voss, Krueger, Henighan, Child, Ramesh, Ziegler, Wu, Winter, Hesse, Chen, Sigler, Litwin, Gray, Chess, Clark, Berner, McCandlish, Radford, Sutskever, and Amodei}]{firstfewshot}
Tom Brown, Benjamin Mann, Nick Ryder, Melanie Subbiah, Jared~D Kaplan, Prafulla Dhariwal, Arvind Neelakantan, Pranav Shyam, Girish Sastry, Amanda Askell, Sandhini Agarwal, Ariel Herbert-Voss, Gretchen Krueger, Tom Henighan, Rewon Child, Aditya Ramesh, Daniel Ziegler, Jeffrey Wu, Clemens Winter, Chris Hesse, Mark Chen, Eric Sigler, Mateusz Litwin, Scott Gray, Benjamin Chess, Jack Clark, Christopher Berner, Sam McCandlish, Alec Radford, Ilya Sutskever, and Dario Amodei. 2020.
\newblock \href {https://proceedings.neurips.cc/paper_files/paper/2020/file/1457c0d6bfcb4967418bfb8ac142f64a-Paper.pdf} {Language models are few-shot learners}.
\newblock In \emph{Advances in Neural Information Processing Systems}, volume~33, pages 1877--1901. Curran Associates, Inc.

\bibitem[{Chen and Gupta(2015)}]{noise_curriculum}
Xinlei Chen and Abhinav Gupta. 2015.
\newblock Webly supervised learning of convolutional networks.
\newblock In \emph{Proceedings of the IEEE international conference on computer vision}, pages 1431--1439.

\bibitem[{Dai et~al.(2023)Dai, Sun, Dong, Hao, Ma, Sui, and Wei}]{understandicl2}
Damai Dai, Yutao Sun, Li~Dong, Yaru Hao, Shuming Ma, Zhifang Sui, and Furu Wei. 2023.
\newblock Why can gpt learn in-context? language models secretly perform gradient descent as meta-optimizers.
\newblock In \emph{Findings of the Association for Computational Linguistics: ACL 2023}, pages 4005--4019.

\bibitem[{Dong et~al.(2023)Dong, Yuan, Lu, Li, Xue, Liu, Wang, Yuan, Zhou, and Zhou}]{sft1}
Guanting Dong, Hongyi Yuan, Keming Lu, Chengpeng Li, Mingfeng Xue, Dayiheng Liu, Wei Wang, Zheng Yuan, Chang Zhou, and Jingren Zhou. 2023.
\newblock How abilities in large language models are affected by supervised fine-tuning data composition.
\newblock \emph{arXiv preprint arXiv:2310.05492}.

\bibitem[{Dong et~al.(2022)Dong, Li, Dai, Zheng, Wu, Chang, Sun, Xu, and Sui}]{defineicl}
Qingxiu Dong, Lei Li, Damai Dai, Ce~Zheng, Zhiyong Wu, Baobao Chang, Xu~Sun, Jingjing Xu, and Zhifang Sui. 2022.
\newblock A survey on in-context learning.
\newblock \emph{arXiv preprint arXiv:2301.00234}.

\bibitem[{Du et~al.(2025{\natexlab{a}})Du, Fang, Li, Li, Jiang, Yu, Guo, Chen, Goh, Tang, He, Liu, and Zhang}]{sft2}
Guodong Du, Zitao Fang, Jing Li, Junlin Li, Runhua Jiang, Shuyang Yu, Yifei Guo, Yangneng Chen, Sim~Kuan Goh, Ho-Kin Tang, Daojing He, Honghai Liu, and Min Zhang. 2025{\natexlab{a}}.
\newblock \href {https://arxiv.org/abs/2505.18713} {Neural parameter search for slimmer fine-tuned models and better transfer}.
\newblock \emph{arXiv preprint arXiv:2505.18713}.

\bibitem[{Du et~al.(2025{\natexlab{b}})Du, Zhou, Li, Li, Shi, Lin, Tang, Li, Liu, Wang, Zhang, and Li}]{rl2}
Guodong Du, Xuanning Zhou, Junlin Li, Zhuo Li, Zesheng Shi, Wanyu Lin, Ho-Kin Tang, Xiucheng Li, Fangming Liu, Wenya Wang, Min Zhang, and Jing Li. 2025{\natexlab{b}}.
\newblock \href {https://arxiv.org/abs/2505.18502} {Knowledge grafting of large language models}.
\newblock \emph{arXiv preprint arXiv:2505.18502}.

\bibitem[{Du et~al.(2023)Du, Watkins, Wang, Colas, Darrell, Abbeel, Gupta, and Andreas}]{rl1}
Yuqing Du, Olivia Watkins, Zihan Wang, C{\'e}dric Colas, Trevor Darrell, Pieter Abbeel, Abhishek Gupta, and Jacob Andreas. 2023.
\newblock Guiding pretraining in reinforcement learning with large language models.
\newblock In \emph{International Conference on Machine Learning}, pages 8657--8677. PMLR.

\bibitem[{Dubey et~al.(2024)Dubey, Jauhri, Pandey, Kadian, Al-Dahle, Letman, Mathur, Schelten, Yang, Fan et~al.}]{llama3}
Abhimanyu Dubey, Abhinav Jauhri, Abhinav Pandey, Abhishek Kadian, Ahmad Al-Dahle, Aiesha Letman, Akhil Mathur, Alan Schelten, Amy Yang, Angela Fan, et~al. 2024.
\newblock The llama 3 herd of models.
\newblock \emph{arXiv preprint arXiv:2407.21783}.

\bibitem[{Geva et~al.(2021)Geva, Khashabi, Segal, Khot, Roth, and Berant}]{strategyqa}
Mor Geva, Daniel Khashabi, Elad Segal, Tushar Khot, Dan Roth, and Jonathan Berant. 2021.
\newblock Did aristotle use a laptop? a question answering benchmark with implicit reasoning strategies.
\newblock \emph{Transactions of the Association for Computational Linguistics}, 9:346--361.

\bibitem[{Gonen et~al.(2023)Gonen, Iyer, Blevins, Smith, and Zettlemoyer}]{perplexity}
Hila Gonen, Srini Iyer, Terra Blevins, Noah~A Smith, and Luke Zettlemoyer. 2023.
\newblock Demystifying prompts in language models via perplexity estimation.
\newblock In \emph{Findings of the Association for Computational Linguistics: EMNLP 2023}, pages 10136--10148.

\bibitem[{Guo et~al.(2025)Guo, Yang, Zhang, Song, Zhang, Xu, Zhu, Ma, Wang, Bi et~al.}]{deepseek}
Daya Guo, Dejian Yang, Haowei Zhang, Junxiao Song, Ruoyu Zhang, Runxin Xu, Qihao Zhu, Shirong Ma, Peiyi Wang, Xiao Bi, et~al. 2025.
\newblock Deepseek-r1: Incentivizing reasoning capability in llms via reinforcement learning.
\newblock \emph{arXiv preprint arXiv:2501.12948}.

\bibitem[{Guo et~al.(2024)Guo, Wang, Wang, Ye, and Zhang}]{relatedwork_icl3}
Qi~Guo, Leiyu Wang, Yidong Wang, Wei Ye, and Shikun Zhang. 2024.
\newblock What makes a good order of examples in in-context learning.
\newblock In \emph{Findings of the Association for Computational Linguistics ACL 2024}, pages 14892--14904.

\bibitem[{Hacohen and Weinshall(2019)}]{curriculum2}
Guy Hacohen and Daphna Weinshall. 2019.
\newblock On the power of curriculum learning in training deep networks.
\newblock In \emph{International conference on machine learning}, pages 2535--2544. PMLR.

\bibitem[{Hongjin et~al.(2022)Hongjin, Kasai, Wu, Shi, Wang, Xin, Zhang, Ostendorf, Zettlemoyer, Smith et~al.}]{votek}
SU~Hongjin, Jungo Kasai, Chen~Henry Wu, Weijia Shi, Tianlu Wang, Jiayi Xin, Rui Zhang, Mari Ostendorf, Luke Zettlemoyer, Noah~A Smith, et~al. 2022.
\newblock Selective annotation makes language models better few-shot learners.
\newblock In \emph{The Eleventh International Conference on Learning Representations}.

\bibitem[{Hou et~al.(2025)Hou, Zhang, Wang, Ma, and Huang}]{complex4}
Yujie Hou, Ting Zhang, Mei Wang, Xuetao Ma, and Hua Huang. 2025.
\newblock \href {https://arxiv.org/abs/2505.16646} {Smart: Self-generating and self-validating multi-dimensional assessment for llms' mathematical problem solving}.
\newblock \emph{arXiv preprint arXiv:2505.16646}.

\bibitem[{Hsieh et~al.(2023)Hsieh, Li, Yeh, Nakhost, Fujii, Ratner, Krishna, Lee, and Pfister}]{smallsft1}
Cheng-Yu Hsieh, Chun-Liang Li, Chih-kuan Yeh, Hootan Nakhost, Yasuhisa Fujii, Alex Ratner, Ranjay Krishna, Chen-Yu Lee, and Tomas Pfister. 2023.
\newblock Distilling step-by-step! outperforming larger language models with less training data and smaller model sizes.
\newblock In \emph{Findings of the Association for Computational Linguistics: ACL 2023}, pages 8003--8017.

\bibitem[{Hu et~al.(2022)Hu, Shen, Wallis, Allen-Zhu, Li, Wang, Wang, and Chen}]{lora}
Edward~J Hu, Yelong Shen, Phillip Wallis, Zeyuan Allen-Zhu, Yuanzhi Li, Shean Wang, Lu~Wang, and Weizhu Chen. 2022.
\newblock \href {https://openreview.net/forum?id=nZeVKeeFYf9} {Lo{RA}: Low-rank adaptation of large language models}.
\newblock In \emph{International Conference on Learning Representations}.

\bibitem[{Kim and Lee(2024)}]{relatedwork_cl1}
Jisu Kim and Juhwan Lee. 2024.
\newblock Strategic data ordering: Enhancing large language model performance through curriculum learning.
\newblock \emph{arXiv preprint arXiv:2405.07490}.

\bibitem[{Koncel-Kedziorski et~al.(2016)Koncel-Kedziorski, Roy, Amini, Kushman, and Hajishirzi}]{mawps}
Rik Koncel-Kedziorski, Subhro Roy, Aida Amini, Nate Kushman, and Hannaneh Hajishirzi. 2016.
\newblock Mawps: A math word problem repository.
\newblock In \emph{Proceedings of the 2016 conference of the north american chapter of the association for computational linguistics: human language technologies}, pages 1152--1157.

\bibitem[{Lee et~al.(2023)Lee, Cho, and Yoo}]{humancurriculum}
Bruce~W Lee, Hyunsoo Cho, and Kang~Min Yoo. 2023.
\newblock Instruction tuning with human curriculum.
\newblock \emph{arXiv preprint arXiv:2310.09518}.

\bibitem[{Levy et~al.(2023)Levy, Bogin, and Berant}]{diverse1}
Itay Levy, Ben Bogin, and Jonathan Berant. 2023.
\newblock Diverse demonstrations improve in-context compositional generalization.
\newblock In \emph{Proceedings of the 61st Annual Meeting of the Association for Computational Linguistics (Volume 1: Long Papers)}, pages 1401--1422.

\bibitem[{Li and Qiu(2023)}]{relatedwork_icl2}
Xiaonan Li and Xipeng Qiu. 2023.
\newblock Finding support examples for in-context learning.
\newblock In \emph{Findings of the Association for Computational Linguistics: EMNLP 2023}, pages 6219--6235.

\bibitem[{Li et~al.(2025{\natexlab{a}})Li, Yang, Li, and Tang}]{icl_future1}
Yanshu Li, JianJiang Yang, Bozheng Li, and Ruixiang Tang. 2025{\natexlab{a}}.
\newblock Cama: Enhancing multimodal in-context learning with context-aware modulated attention.
\newblock \emph{arXiv preprint arXiv:2505.17097}.

\bibitem[{Li et~al.(2025{\natexlab{b}})Li, Yun, Yang, Feng, Huang, and Tang}]{icl_future2}
Yanshu Li, Tian Yun, Jianjiang Yang, Pinyuan Feng, Jinfa Huang, and Ruixiang Tang. 2025{\natexlab{b}}.
\newblock Taco: Enhancing multimodal in-context learning via task mapping-guided sequence configuration.
\newblock \emph{arXiv preprint arXiv:2505.17098}.

\bibitem[{Ling et~al.(2017)Ling, Yogatama, Dyer, and Blunsom}]{aqua}
Wang Ling, Dani Yogatama, Chris Dyer, and Phil Blunsom. 2017.
\newblock Program induction by rationale generation: Learning to solve and explain algebraic word problems.
\newblock In \emph{Proceedings of the 55th Annual Meeting of the Association for Computational Linguistics (Volume 1: Long Papers)}, pages 158--167.

\bibitem[{Liu et~al.(2022)Liu, Shen, Zhang, Dolan, Carin, and Chen}]{kate}
Jiachang Liu, Dinghan Shen, Yizhe Zhang, William~B Dolan, Lawrence Carin, and Weizhu Chen. 2022.
\newblock What makes good in-context examples for gpt-3?
\newblock In \emph{Proceedings of Deep Learning Inside Out (DeeLIO 2022): The 3rd Workshop on Knowledge Extraction and Integration for Deep Learning Architectures}, pages 100--114.

\bibitem[{Liu et~al.(2024)Liu, Liu, Shi, Cheng, and Lu}]{iccl}
Yinpeng Liu, Jiawei Liu, Xiang Shi, Qikai Cheng, and Wei Lu. 2024.
\newblock Let's learn step by step: Enhancing in-context learning ability with curriculum learning.
\newblock \emph{arXiv preprint arXiv:2402.10738}.

\bibitem[{Lu et~al.(2022)Lu, Bartolo, Moore, Riedel, and Stenetorp}]{order}
Yao Lu, Max Bartolo, Alastair Moore, Sebastian Riedel, and Pontus Stenetorp. 2022.
\newblock Fantastically ordered prompts and where to find them: Overcoming few-shot prompt order sensitivity.
\newblock In \emph{Proceedings of the 60th Annual Meeting of the Association for Computational Linguistics (Volume 1: Long Papers)}, pages 8086--8098.

\bibitem[{Margatina et~al.(2023{\natexlab{a}})Margatina, Schick, Aletras, and Dwivedi-Yu}]{perplexity2}
Katerina Margatina, Timo Schick, Nikolaos Aletras, and Jane Dwivedi-Yu. 2023{\natexlab{a}}.
\newblock Active learning principles for in-context learning with large language models.
\newblock In \emph{Findings of the Association for Computational Linguistics: EMNLP 2023}, pages 5011--5034.

\bibitem[{Margatina et~al.(2023{\natexlab{b}})Margatina, Schick, Aletras, and Dwivedi-Yu}]{alicl}
Katerina Margatina, Timo Schick, Nikolaos Aletras, and Jane Dwivedi-Yu. 2023{\natexlab{b}}.
\newblock Active learning principles for in-context learning with large language models.
\newblock In \emph{Findings of the Association for Computational Linguistics: EMNLP 2023}, pages 5011--5034.

\bibitem[{Miao et~al.(2020)Miao, Liang, and Su}]{asdiv}
Shen-yun Miao, Chao-Chun Liang, and Keh-Yih Su. 2020.
\newblock A diverse corpus for evaluating and developing english math word problem solvers.
\newblock In \emph{Proceedings of the 58th Annual Meeting of the Association for Computational Linguistics}, pages 975--984.

\bibitem[{Nguyen and Wong(2023)}]{relatedwork_icl1}
Tai Nguyen and Eric Wong. 2023.
\newblock In-context example selection with influences.
\newblock \emph{arXiv preprint arXiv:2302.11042}.

\bibitem[{Ouyang et~al.(2022)Ouyang, Wu, Jiang, Almeida, Wainwright, Mishkin, Zhang, Agarwal, Slama, Ray et~al.}]{human_tuning}
Long Ouyang, Jeffrey Wu, Xu~Jiang, Diogo Almeida, Carroll Wainwright, Pamela Mishkin, Chong Zhang, Sandhini Agarwal, Katarina Slama, Alex Ray, et~al. 2022.
\newblock Training language models to follow instructions with human feedback.
\newblock \emph{Advances in neural information processing systems}, 35:27730--27744.

\bibitem[{Patel et~al.(2021)Patel, Bhattamishra, and Goyal}]{svamp}
Arkil Patel, Satwik Bhattamishra, and Navin Goyal. 2021.
\newblock Are nlp models really able to solve simple math word problems?
\newblock In \emph{Proceedings of the 2021 Conference of the North American Chapter of the Association for Computational Linguistics: Human Language Technologies}, pages 2080--2094.

\bibitem[{Peng et~al.(2024)Peng, Ding, Yuan, Liu, Zhang, Ouyang, and Tao}]{revisiting}
Keqin Peng, Liang Ding, Yancheng Yuan, Xuebo Liu, Min Zhang, Yuanxin Ouyang, and Dacheng Tao. 2024.
\newblock Revisiting demonstration selection strategies in in-context learning.
\newblock \emph{arXiv preprint arXiv:2401.12087}.

\bibitem[{Platanios et~al.(2019)Platanios, Stretcu, Neubig, Pocz{\'o}s, and Mitchell}]{textlength_curriculum}
Emmanouil~Antonios Platanios, Otilia Stretcu, Graham Neubig, Barnab{\'a}s Pocz{\'o}s, and Tom Mitchell. 2019.
\newblock Competence-based curriculum learning for neural machine translation.
\newblock In \emph{Proceedings of the 2019 Conference of the North American Chapter of the Association for Computational Linguistics: Human Language Technologies, Volume 1 (Long and Short Papers)}, pages 1162--1172.

\bibitem[{Pouransari et~al.(2024)Pouransari, Li, Chang, Vasu, Koc, Shankar, and Tuzel}]{datasetdecomposition}
Hadi Pouransari, Chun-Liang Li, Jen-Hao~Rick Chang, Pavan Kumar~Anasosalu Vasu, Cem Koc, Vaishaal Shankar, and Oncel Tuzel. 2024.
\newblock Dataset decomposition: Faster llm training with variable sequence length curriculum.
\newblock \emph{arXiv preprint arXiv:2405.13226}.

\bibitem[{Qwen et~al.(2025)Qwen, :, Yang, Yang, Zhang, Hui, Zheng, Yu, Li, Liu, Huang, Wei, Lin, Yang, Tu, Zhang, Yang, Yang, Zhou, Lin, Dang, Lu, Bao, Yang, Yu, Li, Xue, Zhang, Zhu, Men, Lin, Li, Tang, Xia, Ren, Ren, Fan, Su, Zhang, Wan, Liu, Cui, Zhang, and Qiu}]{qwen2_5}
Qwen, :, An~Yang, Baosong Yang, Beichen Zhang, Binyuan Hui, Bo~Zheng, Bowen Yu, Chengyuan Li, Dayiheng Liu, Fei Huang, Haoran Wei, Huan Lin, Jian Yang, Jianhong Tu, Jianwei Zhang, Jianxin Yang, Jiaxi Yang, Jingren Zhou, Junyang Lin, Kai Dang, Keming Lu, Keqin Bao, Kexin Yang, Le~Yu, Mei Li, Mingfeng Xue, Pei Zhang, Qin Zhu, Rui Men, Runji Lin, Tianhao Li, Tianyi Tang, Tingyu Xia, Xingzhang Ren, Xuancheng Ren, Yang Fan, Yang Su, Yichang Zhang, Yu~Wan, Yuqiong Liu, Zeyu Cui, Zhenru Zhang, and Zihan Qiu. 2025.
\newblock \href {https://arxiv.org/abs/2412.15115} {Qwen2.5 technical report}.
\newblock \emph{Preprint}, arXiv:2412.15115.

\bibitem[{Robertson et~al.(2009)Robertson, Zaragoza et~al.}]{similarity}
Stephen Robertson, Hugo Zaragoza, et~al. 2009.
\newblock The probabilistic relevance framework: Bm25 and beyond.
\newblock \emph{Foundations and Trends{\textregistered} in Information Retrieval}, 3(4):333--389.

\bibitem[{Rubin et~al.(2022)Rubin, Herzig, and Berant}]{promot_retrieve}
Ohad Rubin, Jonathan Herzig, and Jonathan Berant. 2022.
\newblock Learning to retrieve prompts for in-context learning.
\newblock In \emph{Proceedings of the 2022 Conference of the North American Chapter of the Association for Computational Linguistics: Human Language Technologies}, pages 2655--2671.

\bibitem[{Shi et~al.(2024)Shi, Qing, Yang, Wang, Lei, Lu, Lin, and Li}]{promptso}
Fobo Shi, Peijun Qing, Dong Yang, Nan Wang, Youbo Lei, Haonan Lu, Xiaodong Lin, and Duantengchuan Li. 2024.
\newblock Prompt space optimizing few-shot reasoning success with large language models.
\newblock In \emph{Findings of the Association for Computational Linguistics: NAACL 2024}, pages 1836--1862.

\bibitem[{Soviany et~al.(2020)Soviany, Ardei, Ionescu, and Leordeanu}]{score_curriculum}
Petru Soviany, Claudiu Ardei, Radu~Tudor Ionescu, and Marius Leordeanu. 2020.
\newblock Image difficulty curriculum for generative adversarial networks (cugan).
\newblock In \emph{Proceedings of the IEEE/CVF winter conference on applications of computer vision}, pages 3463--3472.

\bibitem[{Talmor et~al.(2019)Talmor, Herzig, Lourie, and Berant}]{commonsenseqa}
Alon Talmor, Jonathan Herzig, Nicholas Lourie, and Jonathan Berant. 2019.
\newblock Commonsenseqa: A question answering challenge targeting commonsense knowledge.
\newblock In \emph{Proceedings of the 2019 Conference of the North American Chapter of the Association for Computational Linguistics: Human Language Technologies, Volume 1 (Long and Short Papers)}, pages 4149--4158.

\bibitem[{Touvron et~al.(2023)Touvron, Martin, Stone, Albert, Almahairi, Babaei, Bashlykov, Batra, Bhargava, Bhosale et~al.}]{llama2}
Hugo Touvron, Louis Martin, Kevin Stone, Peter Albert, Amjad Almahairi, Yasmine Babaei, Nikolay Bashlykov, Soumya Batra, Prajjwal Bhargava, Shruti Bhosale, et~al. 2023.
\newblock Llama 2: Open foundation and fine-tuned chat models.
\newblock \emph{arXiv preprint arXiv:2307.09288}.

\bibitem[{Wang et~al.(2021)Wang, Chen, and Zhu}]{curriculum_survey}
Xin Wang, Yudong Chen, and Wenwu Zhu. 2021.
\newblock A survey on curriculum learning.
\newblock \emph{IEEE transactions on pattern analysis and machine intelligence}, 44(9):4555--4576.

\bibitem[{Wang et~al.(2024)Wang, Zhou, Chen, and Zhu}]{relatedwork_cl2}
Xin Wang, Yuwei Zhou, Hong Chen, and Wenwu Zhu. 2024.
\newblock Curriculum learning: Theories, approaches, applications, tools, and future directions in the era of large language models.
\newblock In \emph{Companion Proceedings of the ACM on Web Conference 2024}, pages 1306--1310.

\bibitem[{Wei et~al.(2022)Wei, Wang, Schuurmans, Bosma, Xia, Chi, Le, Zhou et~al.}]{cot}
Jason Wei, Xuezhi Wang, Dale Schuurmans, Maarten Bosma, Fei Xia, Ed~Chi, Quoc~V Le, Denny Zhou, et~al. 2022.
\newblock Chain-of-thought prompting elicits reasoning in large language models.
\newblock \emph{Advances in neural information processing systems}, 35:24824--24837.

\bibitem[{Wei et~al.(2016)Wei, Liang, Chen, Shen, Cheng, Feng, Zhao, and Yan}]{image_curriculum}
Yunchao Wei, Xiaodan Liang, Yunpeng Chen, Xiaohui Shen, Ming-Ming Cheng, Jiashi Feng, Yao Zhao, and Shuicheng Yan. 2016.
\newblock Stc: A simple to complex framework for weakly-supervised semantic segmentation.
\newblock \emph{IEEE transactions on pattern analysis and machine intelligence}, 39(11):2314--2320.

\bibitem[{Wies et~al.(2024)Wies, Levine, and Shashua}]{iclsurvey1}
Noam Wies, Yoav Levine, and Amnon Shashua. 2024.
\newblock The learnability of in-context learning.
\newblock \emph{Advances in Neural Information Processing Systems}, 36.

\bibitem[{Wolfson et~al.(2020)Wolfson, Geva, Gupta, Gardner, Goldberg, Deutch, and Berant}]{qdmr}
Tomer Wolfson, Mor Geva, Ankit Gupta, Matt Gardner, Yoav Goldberg, Daniel Deutch, and Jonathan Berant. 2020.
\newblock Break it down: A question understanding benchmark.
\newblock \emph{Transactions of the Association for Computational Linguistics}, 8:183--198.

\bibitem[{Wu et~al.(2023{\natexlab{a}})Wu, Wang, Ye, and Kong}]{similarity2}
Zhiyong Wu, Yaoxiang Wang, Jiacheng Ye, and Lingpeng Kong. 2023{\natexlab{a}}.
\newblock Self-adaptive in-context learning: An information compression perspective for in-context example selection and ordering.
\newblock In \emph{Proceedings of the 61st Annual Meeting of the Association for Computational Linguistics (Volume 1: Long Papers)}, pages 1423--1436.

\bibitem[{Wu et~al.(2023{\natexlab{b}})Wu, Wang, Ye, and Kong}]{saicl}
Zhiyong Wu, Yaoxiang Wang, Jiacheng Ye, and Lingpeng Kong. 2023{\natexlab{b}}.
\newblock Self-adaptive in-context learning: An information compression perspective for in-context example selection and ordering.
\newblock In \emph{Proceedings of the 61st Annual Meeting of the Association for Computational Linguistics (Volume 1: Long Papers)}, pages 1423--1436.

\bibitem[{Xie et~al.(2024)Xie, Luo, Stern, Du, and Cheng}]{diverse2}
Shan Xie, Man Luo, Chadly~Daniel Stern, Mengnan Du, and Lu~Cheng. 2024.
\newblock Demoshapley: Valuation of demonstrations for in-context learning.
\newblock \emph{arXiv preprint arXiv:2410.07523}.

\bibitem[{Xu et~al.(2024)Xu, Liu, Pasupat, Kazemi et~al.}]{iclsurvey2}
Xin Xu, Yue Liu, Panupong Pasupat, Mehran Kazemi, et~al. 2024.
\newblock In-context learning with retrieved demonstrations for language models: A survey.
\newblock \emph{arXiv preprint arXiv:2401.11624}.

\bibitem[{Yao et~al.(2024)Yao, Yu, Zhao, Shafran, Griffiths, Cao, and Narasimhan}]{cot_future1}
Shunyu Yao, Dian Yu, Jeffrey Zhao, Izhak Shafran, Tom Griffiths, Yuan Cao, and Karthik Narasimhan. 2024.
\newblock Tree of thoughts: Deliberate problem solving with large language models.
\newblock \emph{Advances in Neural Information Processing Systems}, 36.

\bibitem[{Ye et~al.(2023)Ye, Chen, Xu, Zu, Shao, Liu, Cui, Zhou, Gong, Shen et~al.}]{gpt3}
Junjie Ye, Xuanting Chen, Nuo Xu, Can Zu, Zekai Shao, Shichun Liu, Yuhan Cui, Zeyang Zhou, Chao Gong, Yang Shen, et~al. 2023.
\newblock A comprehensive capability analysis of gpt-3 and gpt-3.5 series models.
\newblock \emph{arXiv preprint arXiv:2303.10420}.

\bibitem[{Ye et~al.(2025)Ye, Huang, Xiao, Chern, Xia, and Liu}]{lessismore}
Yixin Ye, Zhen Huang, Yang Xiao, Ethan Chern, Shijie Xia, and Pengfei Liu. 2025.
\newblock Limo: Less is more for reasoning.
\newblock \emph{arXiv preprint arXiv:2502.03387}.

\bibitem[{Zhang et~al.(2023)Zhang, Yang, Yuan, and Yao}]{complex2}
Yifan Zhang, Jingqin Yang, Yang Yuan, and Andrew Chi-Chih Yao. 2023.
\newblock Cumulative reasoning with large language models.
\newblock \emph{arXiv preprint arXiv:2308.04371}.

\bibitem[{Zhang et~al.(2022)Zhang, Zhang, Li, and Smola}]{autocot}
Zhuosheng Zhang, Aston Zhang, Mu~Li, and Alex Smola. 2022.
\newblock Automatic chain of thought prompting in large language models.
\newblock \emph{arXiv preprint arXiv:2210.03493}.

\bibitem[{Zhao and Zhang(2024)}]{complex3}
Jinman Zhao and Xueyan Zhang. 2024.
\newblock \href {https://openreview.net/forum?id=wLQ3I0F1oj} {Large language model is not a (multilingual) compositional relation reasoner}.
\newblock In \emph{First Conference on Language Modeling}.

\bibitem[{Zheng et~al.(2024)Zheng, Zhang, Zhang, YeYanhan, and Luo}]{llamafactory}
Yaowei Zheng, Richong Zhang, Junhao Zhang, YeYanhan YeYanhan, and Zheyan Luo. 2024.
\newblock \href {https://doi.org/10.18653/v1/2024.acl-demos.38} {{L}lama{F}actory: Unified efficient fine-tuning of 100+ language models}.
\newblock In \emph{Proceedings of the 62nd Annual Meeting of the Association for Computational Linguistics (Volume 3: System Demonstrations)}, pages 400--410, Bangkok, Thailand. Association for Computational Linguistics.

\end{thebibliography}

\appendix
\section{BREAK Dataset Description}\label{append1}

BREAK is a dataset proposed by the Allen Institute~\cite{qdmr}.
This work introduces the Question Decomposition Meaning Representation (QDMR), which breaks down a question into several sub-questions for solving and represents it as a sequence of steps.
The dataset collects 60,150 question and QDMR pairs from several public datasets. To represent various questions as a unified sequence of steps, they customized 13 types of operations, converting the solution process for all questions into sequences of these operations.
The specific operations and their templates are shown in Table~\ref{operators}.
The decomposition and formalization of questions can be found in Figure~\ref{psl_and_icl} and Figure~\ref{qdmr}.
Table~\ref{operator_prevalence} shows the distribution of operations in the BREAK dataset, that is, the proportion of each operation appearing in a single data point.
Table 6 shows the distribution of the total number of sub-questions after decomposition in the dataset.

Based on the BREAK dataset, we constructed an instruction set to analyze the problem-solving logic. Specific examples and explanations of the instruction set are provided in Table~\ref{instructionset}.

\section{Fine-Tuning Details}\label{finetune}

We performed LoRA fine-tuning on the Llama3-8B model using the aforementioned instruction set. The specific hyperparameters are as follows:
the \texttt{cutoff\_len} is set to 1024, the learning rate is set to $5 \times 10^{-5}$, the fine-tuning parameters are specified as \texttt{all}, \texttt{lora\_rank} is set to 8, \texttt{lora\_alpha} is set to 16, the optimizer used is AdamW, the model is trained for 4 epochs, and the best model is selected based on the BLEU score.

\section{Prompt Template}\label{append2}
Table~\ref{break_prompt} shows the prompt templates used for fine-tuning problem-solving logic analysis.

\noindent
Table~\ref{svamp_prompt}--\ref{sqa_prompt} shows the full prompt example for in-context learning on the different benchmarks.

\section{Supplementary Details}\label{append3}
Our experiments utilized the llama-factory~\cite{llamafactory} project, which includes model fine-tuning and in-context learning. The CPU used in the experiments is an Intel(R) Xeon(R) Platinum 8358 CPU @ 2.60GHz, and the GPU is an NVIDIA Tesla A800 80G. The hyperparameters were set according to the default configuration file provided by llama-factory. The prompt length was set to 4096, and the maximum answer output length was set to 1024. To ensure output stability, the temperature was set to 0.01.
In our study, we used ChatGPT to assist in coding.

\begin{table*}[t!]
  \begin{center}
  \scriptsize
  \begin{tabular}{p{1.5cm} p{4cm} p{4.5cm} p{4.5cm}}
  \hline \bf Operator & \bf Template / Signature & \bf Question & \bf Decomposition  \\ \hline
  \bf Select & Return [entities] \newline $\texttt{w} \rightarrow \texttt{S$_\texttt{e}$}$ & How many touchdowns were scored overall? &  1. Return touchdowns \newline 2. Return the number of \#1 \\ \hline
  \bf Filter & Return [ref] [condition] \newline $\texttt{S$_\texttt{o}$}\texttt{,} \texttt{w} \rightarrow \texttt{S$_\texttt{o}$}$ & I would like a flight from Toronto to San Diego please. & 1. Return flights \newline 2. Return \#1 from Toronto \newline 3. Return \#2 to San Diego \\ \hline
  \bf Project & Return [relation] of [ref] \newline $\texttt{w}\texttt{,} \texttt{S$_\texttt{e}$} \rightarrow \texttt{S$_\texttt{o}$}$ & Who is the head coach of the Los Angeles Lakers? & 1. Return the Los Angeles Lakers \newline 2. Return the head coach of \#1 \\ \hline
  \bf Aggregate & Return [aggregate] of [ref] \newline $\texttt{w}_{\texttt{agg}}\texttt{,} \texttt{S$_\texttt{o}$} \rightarrow \texttt{n}$ & How many states border Colorado? & 1. Return Colorado  \newline 2. Return border states of \#1 \newline 3. Return the number of \#2 \\ \hline
  \bf Group & Return [aggregate] [ref1] for each [ref2] \newline $\texttt{w}_{\texttt{agg}}\texttt{,} \texttt{S$_\texttt{o}$,} \texttt{S$_\texttt{e}$} \rightarrow \texttt{S$_\texttt{n}$}$ & How many female students are there in each club? & 1. Return clubs \newline
  2. Return female students of \#1 \newline
  3. Return the number of \#2 for each \#1  \\ \hline
  \bf Superlative & Return [ref1] where [ref2] is [highest / lowest] \newline $\texttt{S$_\texttt{e}$,} \texttt{S$_\texttt{n}$,} \texttt{w}_{\texttt{sup}} \rightarrow \texttt{S$_\texttt{e}$}$ & What is the keyword, which has been contained by the most number of papers?  & 1. Return papers \newline 
  2. Return keywords of \#1 \newline
  3. Return the number of \#1 for each \#2 \newline
  4. Return \#2 where \#3 is highest \\ \hline
  \bf Comparative & Return [ref1] where [ref2] [comparison] [number] \newline $\texttt{S$_\texttt{e}$,} \texttt{S$_\texttt{n}$,} \texttt{w}_{\texttt{com}}\texttt{,} \texttt{n} \rightarrow \texttt{S$_\texttt{e}$}$ & Who are the authors who have more than 500 papers? & 1. Return authors \newline
  2. Return papers of \#1 \newline
  3. Return the number of \#2 for each of \#1 \newline
  4. Return \#1 where \#3 is more than 500 \\ \hline
  \bf Union & Return [ref1] , [ref2] \newline $\texttt{S$_\texttt{o}$,} \texttt{S$_\texttt{o}$} \rightarrow \texttt{S$_\texttt{o}$}$ & Tell me who the president and vice-president are? & 1. Return the president \newline
  2. Return the vice-president \newline
  3. Return \#1 , \#2  \\ \hline
  \bf Intersection & Return [relation] in both [ref1] and [ref2] \newline $\texttt{w,} \texttt{S$_\texttt{e}$,} \texttt{S$_\texttt{e}$} \rightarrow \texttt{S$_\texttt{o}$}$ & Show the parties that have representatives in both New York state and representatives in Pennsylvania state. & 1. Return representatives \newline 2. Return \#1 in New York state \newline 3. Return \#1 in Pennsylvania state \newline 4. Return parties in  both \#2 and \#3 \\ \hline
  \bf Discard & Return [ref1] besides [ref2] \newline $\texttt{S$_\texttt{o}$,} \texttt{S$_\texttt{o}$} \rightarrow \texttt{S$_\texttt{o}$}$ & Find the professors who are not playing Canoeing. & 1. Return professors \newline 2. Return \#1 playing Canoeing \newline 3. Return \#1 besides \#2 \\ \hline
  \bf Sort & Return [ref1] sorted by [ref2] \newline $\texttt{S$_\texttt{e}$,} \texttt{S$_\texttt{n}$} \rightarrow \texttt{$\langle \texttt{e}_1...\texttt{e}_k \rangle$}$ & Find all information about student addresses, and sort by monthly rental. & 1. Return students \newline 2. Return addresses of \#1 \newline 3. Return monthly rental of  \#2 \newline 4. Return \#2 sorted by \#3
   \\ \hline
  \bf Boolean & Return [if / is] [ref1] [condition] [ref2] \newline $\texttt{S$_\texttt{o}$,} \texttt{w,} \texttt{S$_\texttt{o}$} \rightarrow \texttt{b}$ & Were Scott Derrickson and Ed Wood of the same nationality? & ... \newline 3. Return the nationality of \#1 \newline 4. Return the nationality of \#2 \newline 5. Return if \#3 is the same as \#4 \\ \hline
  \bf Arithmetic & Return the [arithmetic] of [ref1] and [ref2] \newline $\texttt{w}_{\texttt{ari}}\texttt{,} \texttt{n,} \texttt{n} \rightarrow \texttt{n}$ & How many more red objects are there than blue objects? & ... \newline 3. Return the number of \#1 \newline 4. Return the number of \#2 \newline 5. Return the difference of \#3 and \#4 \\ \hline
  \end{tabular}
  \end{center}
  \caption{\label{operators} The 13 operator types of QDMR steps. Listed are, the natural language template used to express the operator, the operator signature and an example question that uses the query operator in its decomposition.}
  \end{table*}

  \begin{table}[t]
      \begin{center}
      \normalsize
      \begin{tabular}{cc}
      \hline \bf Operator & \bf QDMR \\ \hline
      \bf\texttt{SELECT} & 100\% \\ 
      \bf\texttt{PROJECT} & 69.0\% \\ 
      \bf\texttt{FILTER} & 53.2\%  \\ 
      \bf\texttt{AGGREGATE} & 38.1\%  \\ 
      \bf\texttt{BOOLEAN} & 30.0\% \\ 
      \bf\texttt{COMPARATIVE} & 17.0\% \\ 
      \bf\texttt{GROUP} & 9.7\%  \\ 
      \bf\texttt{SUPERLATIVE} & 6.3\% \\ 
      \bf\texttt{UNION} & 5.5\% \\ 
      \bf\texttt{ARITHMETIC} & 5.4\% \\ 
      \bf\texttt{DISCARD} & 3.2\% \\ 
      \bf\texttt{INTERSECTION} & 2.7\% \\ 
      \bf\texttt{SORT} & 0.9\% \\ \hline
      Total & 60,150 \\ \hline
      \end{tabular}
      \end{center}
      \caption{\label{operator_prevalence} 
      Operator prevalence in BREAK, that is, the proportion of each operator appearing in a single data point.
      }
  \end{table}
  
  \begin{table}[t]
    \begin{center}
    \normalsize
    \begin{tabular}{cc}
    \hline \bf Steps & \bf QDMR \\ \hline
    1-2 & 10.7\% \\ 
    3-4 & 44.9\% \\ 
    5-6 & 27.0\% \\ 
    7-8 & 10.1\% \\ 
    9+ & 7.4\% \\ \hline
    
    \end{tabular}
    \end{center}
    \caption{\label{qdmr_lengths} 
    The distribution of the total number of QDMR sub-questions.
  }
  \end{table}

  \begin{table*}[!ht]
    \centering
    \begin{tabular}{p{13cm}}
    \hline
    \textbf{Input} \\ 
    \hline
    \noindent\textit{\textbackslash{}\textbackslash{}The input is a problem to be solved, such as:}\\
    what flights are available tomorrow from denver to philadelphia? \\ 
    \hline   
    \textbf{Label} \\ 
    \hline
    \textit{
    \noindent\textbackslash{}\textbackslash{}
    The label contains <operator> and <formal language>. \newline
    \textbackslash{}\textbackslash{}
    <operator> is an ordered set composed of the aforementioned custom operations. \newline
    \textbackslash{}\textbackslash{}
    <formal language> is the formalized language that provides a detailed description of each operator.
    }

    \textless operators\textgreater: ['select', 'filter', 'filter', 'filter'] \newline \textless formal language\textgreater: ["SELECT['flights']", "FILTER['\#1', 'from denver']", "FILTER['\#2', 'to philadelphia']", "FILTER['\#3', 'if available']"] \\ 
    \hline
    \end{tabular}
    \caption{\label{instructionset}
    Examples and Explanation of Instruction Sets Based on the BREAK Dataset
}
\end{table*}

  \begin{table*}[!ht]
    \centering
    \begin{tabular}{p{13cm}}
    \hline
    \textbf{Prompt} \\ 
    \hline
    You are a helpful assistant. Please break down in order the operations \textless operations\textgreater{} required to solve the following problems, and the process of solving the problem according to the operations \textless programs\textgreater{}: \newline what flights are available tomorrow from denver to philadelphia? \\ 
    \hline   
    \textbf{Label} \\ 
    \hline
    \textless operators\textgreater: ['select', 'filter', 'filter', 'filter'] \newline \textless formal language\textgreater: ["SELECT['flights']", "FILTER['\#1', 'from denver']", "FILTER['\#2', 'to philadelphia']", "FILTER['\#3', 'if available']"] \\ 
    \hline
    \end{tabular}
    \caption{\label{break_prompt}
    Fine-tuning prompts for problem-solving logic analysis}
\end{table*}

\begin{table*}[!ht]
  \centering
  \begin{tabular}{p{15cm}}
  \textbf{Prompt} \\ 
  \hline
  \textbf{System prompt}\\
  Please provide the answer in the following format: "The final answer is <answer>" \\
  \textbf{User input} \\
  question: Being his favorite, he saved checking on the grapevines for his last stop. He was told by 235 of the pickers that they fill 100 drums of raspberries per day and 221 drums of grapes per day. How many drums of grapes would be filled in 77 days? \\
  answer: Equation is ( 221.0 * 77.0 ). The final answer is 17017.0 \\
  \\
  question: Tiffany was collecting cans for recycling. On Monday she had 4 bags of cans. The next day she found some more bags worth of cans. If she had a total of 6 bags altogether, how many bags did she find on the next day? \\
  answer: Equation is ( 6.0 - 4.0 ). The final answer is 2.0 \\
  \\
  question: After a typhoon, 13 trees in Haley's backyard died. If she had grown 3 trees initially, how many more trees died in the typhoon than those that survived? \\
  answer: Equation is ( 13.0 - ( 3.0 - 13.0 ) ). The final answer is 23.0 \\
  \\
  question: Brenda's mother made cookies for 5 people. She prepared 22 cookies but had to throw away 17 cookies. If each of them had the same number of cookies, how many did each of them have? \\
  answer: Equation is ( ( 22.0 - 17.0 ) / 5.0 ). The final answer is 1.0 \\
  \\
  question: Haley grew 9 trees in her backyard. After a typhoon 4 died. Then she grew 5 more trees. How many trees does she have left? \\

  \hline
  \textbf{Number of problem-solving logic step:} \\4->5->6->7->7 \\
  \hline

  \end{tabular}
  \caption{\label{svamp_prompt}
  Full prompt example for in-context learning on the SVAMP dataset}
\end{table*}

\begin{table*}[!ht]
  \centering
  \begin{tabular}{p{15cm}}
  \textbf{Prompt} \\ 
  \hline
  \textbf{System prompt}\\
  Please provide the answer in the following format: "The final answer is <answer>" \\
  \textbf{User input} \\
  question: What is the probability of rolling one six-sided dice, and getting a different number on each die?  A)1/12 B)1/3 C)1 D)5/9 E)7/18 \\
  answer: Let's think step by step. MAGOOSHOFFICIAL SOLUTION. For the first die, we can roll any one of six numbers. For the second die, we can roll any number save for the number we rolled on the first die, giving us 5 possibilities. For the third die, we can roll four different numbers (we can’t roll the number we rolled on the first or second die). 6 possibilities out of 216 total possibilities. 6/6 = 1. The final answer is C. \\
  \\
  question: If q is the square of a positive integer, which of the following must be equal to the square of the next positive integer? A)\(\sqrt{n + 1}\) B)\(n + 1\) C)\(n^2 + 1\) D)\(q + 2\sqrt{q} + 1\) E)\(n^2 + 2n + 1\) \\
  answer: Let's think step by step. If q is the square of a positive integer, then \(q = (x)^2\) where x is a positive integer. To calculate, \((x+1)^2 = x^2 + 2x + 1\), which is \(q + 2\sqrt{q} + 1\). This should be D. The final answer is D. \\
  \\
  question: If Tim had lunch at \$50 and he gave 20\% tip, how much did he spend? A)\$60.00 B)\$35.42 C)\$60.60 D)\$21.56 E)\$78.45 \\
  answer: Let's think step by step. The tip is 20\% of what he paid for lunch. Tip = 20\% of 50.00 = \$10.00. Total spent = 50.00 + 10.00 = \$60.00. The final answer is A. \\
  \\
  question: Carl is facing very difficult financial times and can only pay the interest on a \$10,000 loan he has taken. The bank charges him a quarterly compound rate of 4\%. What is the approximate interest he pays annually? A)\$1600 B)\$2000 C)\$2150 D)\$2500 E)\$12000 \\
  answer: Let's think step by step. The bank charges a 4\% quarterly compounded annual rate. Per quarter rate is (16/4)\% = 4\%. Thus, the quarterly compounded interest will be slightly more than \$1600. The final answer is A. \\
  \\
  question: A shopkeeper employed a servant at a monthly salary of 1500. In addition to it, he agreed to pay him a commission of 15\% on the monthly sale. How much sale in Rupees should the servant do if he wants his monthly income as 6000?  A)30000 B)415000 C)31500 D)50000 E)None of these \\

  \hline
  \textbf{Number of problem-solving logic step:} \\2->3->4->5->6 \\
  \hline

  \end{tabular}
  \caption{\label{aqua_prompt}
  Full prompt example for in-context learning on the AQuA dataset}
\end{table*}

\begin{table*}[!ht]
  \centering
  \begin{tabular}{p{15cm}}
  \textbf{Prompt} \\ 
  \hline
  \textbf{System prompt}\\
  Please provide the answer in the following format: "The final answer is <answer>" \\
  \textbf{User input} \\
  question: A shopkeeper bought 150 packets of milk. Each packet contained 250 ml of milk. If one fluid ounce is equal to 30 ml, how many ounces of milk did he buy?\\
  nanswer: Let's think step by step. If the shopkeeper bought 150 packets of milk, each packet containing 250ml of milk, all the packets had a total of 250*150 =<<150*250=37500>>37500ml.Since one ounce equal 30 ml, the total amount of milk that the shopkeeper bought in oz is 37500/30=<<37500/30=1250>>1250 oz of milk. The final answer is 1250\\
  \\
  question: Twenty gallons of tea were poured into 80 containers. Geraldo drank 3.5 containers. How many pints of tea did Geraldo drink?\\
  answer: Let's think step by step. 20 gallons = 160 pints. 160/80 = <<160/80=2>>2 pints.3.5 * 2 pints = <<3.5*2=7>>7 pints. Geraldo drank 7 pints of tea. The final answer is 7\\
  \\
  question: During the holidays, Lance works as a merchandiser. He works 35 hours a week, spread equally over 5 workdays. If Lance earns \$9 an hour, how much does he make on each workday?\\
  answer: Let's think step by step. Lance works 35 / 5 = <<35/5=7>>7 hours a day. So he makes \$9 x 7 = \$<<9*7=63>>63 on each workday. The final answer is 63\\
  \\  
  question: A snack machine accepts only quarters. Candy bars cost ¢25, each piece of chocolate costs ¢75, and a pack of juice costs ¢50. How many quarters are needed to buy three candy bars, two pieces of chocolate, and one pack of juice?\\
  \\
  answer: Let's think step by step. Three candy bars cost ¢25 x 3 = ¢<<25*3=75>>75. Two pieces of chocolate cost ¢75 x 2 = ¢<<75*2=150>>150. So, the total amount needed to buy those is ¢75 + ¢150 + ¢50 = ¢<<75+150+50=275>>275. Since a quarter is equal to ¢25, therefore ¢275/¢25 = <<275/25=11>>11 quarters are needed. The final answer is 11\\
  \\
  question: Mark makes custom dog beds. A bed for a Rottweiler takes 8 pounds of stuffing, a bed for a chihuahua takes 2 pounds of stuffing, and a bed for a collie takes the average amount of stuffing between the first two kinds of beds. How many pounds of stuffing does Mark need to make 4 chihuahua beds and 3 collie beds?\\

  \hline
  \textbf{Number of problem-solving logic step:} \\5->6->7->8->8 \\
  \hline

  \end{tabular}
  \caption{\label{gsm8k_prompt}
  Full prompt example for in-context learning on the Gsm8k dataset}
\end{table*}

\begin{table*}[!ht]
  \centering
  \begin{tabular}{p{15cm}}
  \textbf{Prompt} \\ 
  \hline
  \textbf{System prompt}\\
  Please provide the answer in the following format: "The final answer is <answer>" \\
  \textbf{User input} \\
  question: What is the only was to recover from exhaustion? A. mediate B. have rest C. stay in bed D. run out of steam E. go to sleep\\
  answer: B\\
  \\
  question: Google Maps and other highway and street GPS services have replaced what? A. united states B. mexico C. countryside D. atlas E. oceans\\
  answer: D\\
  \\
  question: You can share files with someone if you have a connection to a what? A. freeway B. radio C. wires D. computer network E. electrical circuit\\
  answer: D\\
  \\
  question: If a person isn't able to pay their bills what must they do? A. know everything B. acknowledgment C. make more money D. throw a party E. spare time\\

  \hline
  \textbf{Number of problem-solving logic step:} \\1->2->3->3 \\
  \hline

  \end{tabular}
  \caption{\label{comqa_prompt} Full prompt example for in-context learning on the ComSenQA dataset}
\end{table*}

\begin{table*}[!ht]
  \centering
  \begin{tabular}{p{15cm}}
  \textbf{Prompt} \\ 
  \hline
  \textbf{System prompt}\\
  Please provide the answer in the following format: "The final answer is yes or no" \\
  \textbf{User input} \\
  question: Can you buy Casio products at Petco?\\
  answer: Casio is a manufacturer of consumer electronics and watches. Petco is a chain store that sells pet supplies like food, bowls, litter, toys, cages and grooming equipment. The final answer is no\\
  \\
  question: Did Clark Gable appear in any movies scored by John Williams?\\
  answer: Clark Gable died in 1960. John Williams scored his first movie in 1961. The final answer is no\\
  \\
  question: Could a dandelion suffer from hepatitis?\\
  answer: Only creatures that contain a liver can suffer from hepatitis. The liver is an organ only found in vertebrates. Vertebrates exist in the kingdom Animalia. Dandelions are plants in the kingdom Plantae. The final answer is no\\
  \\
  question: Did Mozart ever buy anything from Dolce \& Gabbana?\\
  \hline
  \textbf{Number of problem-solving logic step:} \\ 2->3->4->4 \\
  \hline

  \end{tabular}
  \caption{\label{sqa_prompt}
  Full prompt example for in-context learning on the StrategyQA dataset}
\end{table*}

\end{document}